%% file: main.tex
\documentclass[runningheads]{llncs}

\usepackage{eccv}

\usepackage{eccvabbrv}

\usepackage[accsupp]{axessibility}  

\input{math_commands.tex}

\usepackage{hyperref}
\usepackage{url}
\usepackage{multirow} 
\usepackage{wrapfig}
\usepackage{subcaption}
\usepackage{caption} 
\usepackage{mathtools}
\usepackage{amsmath}
\usepackage{algorithm,algorithmic}
\usepackage{xcolor}
\usepackage{graphicx}
\usepackage{booktabs}
\usepackage{thmtools, thm-restate}

\definecolor{expgreen}{RGB}{8, 155, 100}
\definecolor{expgray}{RGB}{180, 180, 180}
\definecolor{darkgreen}{RGB}{0,128,0}

\definecolor{citecolor}{HTML}{0071BC}
\definecolor{linkcolor}{HTML}{ED1C24}
\hypersetup{
colorlinks=true,
linkcolor=linkcolor,
citecolor=citecolor,
}

\usepackage{orcidlink}

\begin{document}

\title{Continuous Speculative Decoding for Autoregressive Image Generation} 

\titlerunning{Continuous Speculative Decoding for Autoregressive Image Generation}

\author{Zili Wang\inst{1,2}\orcidlink{0009-0003-0455-7631} \and
Zheng Zhang\inst{3}\orcidlink{0009-0002-6391-4814} \and
Kun Ding\inst{1,2}\orcidlink{0000-0002-2256-8815} \and
Qi Yang\inst{1,2}\orcidlink{0000-0001-8373-6096} \and \\
Fei Li\inst{4}\orcidlink{0000-0002-0400-678X} \and
Shiming Xiang\inst{1,2}\thanks{Corresponding author.}\orcidlink{0000-0002-2089-9733}
}

\authorrunning{Z. Wang et al.}

\institute{School of Artificial Intelligence, University of Chinese Academy of Sciences, China \and
MAIS, Institute of Automation, Chinese Academy of Sciences, China
\email{smxiang@nlpr.ia.ac.cn} \and
JD.COM Inc, China \and China Tower Corporation Limited, China
}

\maketitle

\input{sec/0_abstract}

\input{sec/1_intro}

\input{sec/2_related_work}
\input{figure_tex/figure_main}
\input{sec/3_methodology}
\input{sec/4_experiment}

\input{sec/5_conclusion}

%
%
\bibliographystyle{splncs04}
\bibliography{main}

\clearpage
\appendix
\input{sec/X_suppl}

\end{document}

%% file: math_commands.tex
\usepackage{amsmath,amsfonts,bm}

\def\Figref#1{Fig.~\ref{#1}}

\def\eqref#1{equation~\ref{#1}}
\def\Eqref#1{Equation~\ref{#1}}

\def\Algref#1{Algorithm~\ref{#1}}

\def\1{\bm{1}}

\DeclareMathAlphabet{\mathsfit}{\encodingdefault}{\sfdefault}{m}{sl}
\SetMathAlphabet{\mathsfit}{bold}{\encodingdefault}{\sfdefault}{bx}{n}



%% file: sec/0_abstract.tex
\input{figure_tex/figure_teaser}
\begin{abstract}
Continuous visual autoregressive (AR) models have demonstrated promising performance in image generation, but their inherently sequential nature results in slow inference speed. Speculative decoding, a successful acceleration technique for large language models (LLMs), has effectively accelerated discrete visual AR models. However, the absence of an analogous theory for continuous distributions precludes its use in accelerating continuous AR models.
To fill this gap, this work presents continuous speculative decoding, and addresses challenges from: 1) low acceptance rate, caused by inconsistent output distribution modeled by target and draft models, and 2) modified distribution without analytic expression, caused by a complex integral. For challenge 1), we address low acceptance rates through an approximated criterion, a novel denoising trajectory alignment strategy based on reparameterization proximity, and token pre-filling. For challenge 2), we introduce acceptance-rejection sampling algorithm with an appropriate upper bound, thereby avoiding explicitly calculating the integral. Furthermore, our denoising trajectory alignment is also reused in acceptance-rejection sampling, effectively avoiding repetitive diffusion model inference.
Extensive experiments on various models at 256×256 and 512×512 resolutions demonstrate that our approach achieves over 2× wall-time speedup while preserving the image generation quality.
\keywords{Visual autoregressive models \and Image generation \and Speculative decoding \and Diffusion models}
\end{abstract}

%% file: figure_tex/figure_teaser.tex

\begin{figure}[h]
    \centering
    \includegraphics[width=1.0\linewidth]{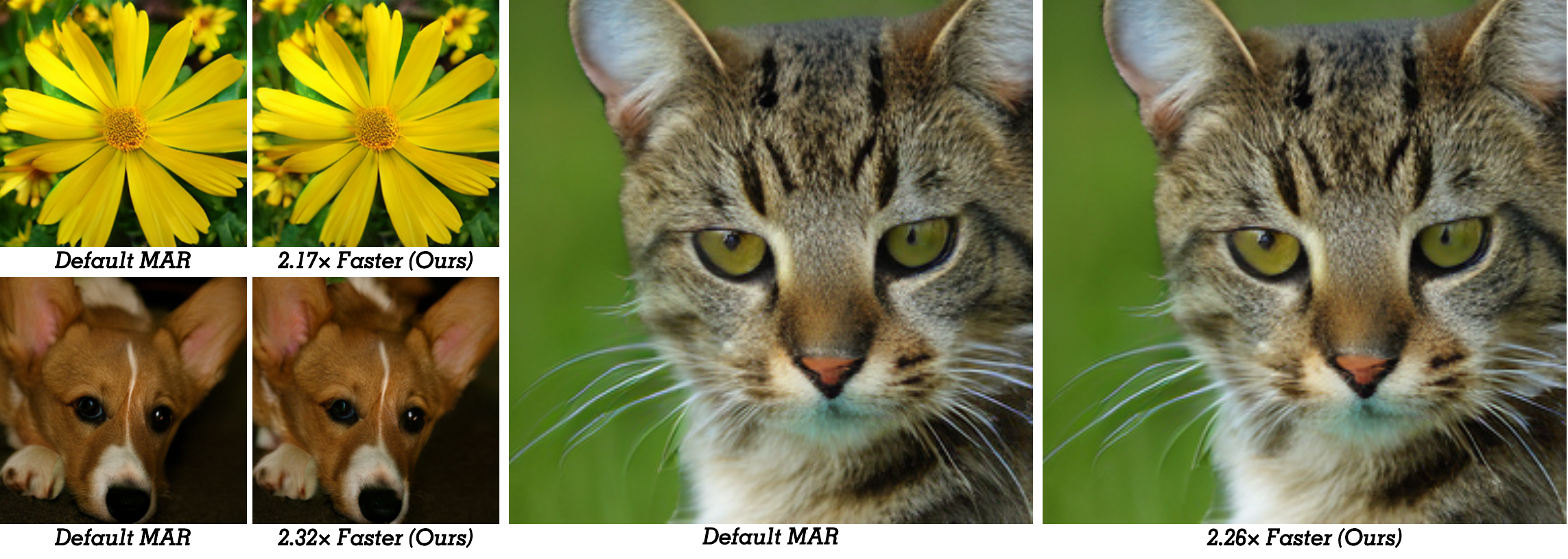}
    \caption{
    Continuous speculative decoding accelerates the inference speed while maintaining the generation quality. For each panel, left: image generated by MAR; right: image generated by MAR with continuous speculative decoding with speed-up ratio. 
    } 
    \label{fig:teaser}
    \vspace{-2em}
\end{figure}

%% file: sec/1_intro.tex
\section{Introduction}\label{sec:intro}
Autoregressive (AR) models have demonstrated significant potential and achieved competitive performance in image generation tasks~\cite{van2016pixelrnn,van2016pixelcnn,esser2021taming,yu2023magvit, VAR}. Existing approaches operate either in discrete token spaces using vector quantization~\cite{yu2023magvit,mentzer2023finite} or in continuous latent spaces with a diffusion model~\cite{tschannen2025givt,li2024mar,xu2024discodiff,ren2025xar}. Discrete AR models suffer from training instability and reconstruction degradation introduced by vector quantization~\cite{yu2023magvit,mentzer2023finite}. In contrast, continuous visual AR models predict continuous latent variables via denoising-based transitions conditioned on previous tokens~\cite{ho2020ddpm,nichol202ddim}, overcome the issues of the discrete counterparts and offer a promising solution for autoregressive image generation. However, similar to LLMs, continuous visual AR models remain slow at inference due to their inherently sequential decoding process.
\input{figure_tex/figure_intro}

To accelerate autoregressive inference, speculative decoding~\cite{leviathan2023fast,chen2023acc} has proven effective for LLMs. This algorithm employs a draft-and-verification mechanism, where a smaller draft model generates several draft tokens, and which are then verified in parallel by a larger and more powerful target model. The acceptance of each draft token depends on the probability ratio (likelihood ratio) between the target and draft model distributions. Tokens with higher ratios are more likely to be accepted. Recent work has extended speculative decoding to discrete visual AR models~\cite{jang2024lantern,teng2024sjd}. However, these methods assume discrete categorical output distributions and do not directly extend to continuous domains.

This paper introduces \textbf{Continuous Speculative Decoding}, a novel framework to accelerate inference for continuous visual AR models. Applying speculative decoding to continuous distributions presents two core challenges, as illustrated in \Figref{fig:figure_intro}:
a) \textbf{Low acceptance rate.} the draft and target models often learn divergent data distributions. As a result, samples generated by the draft model may lie outside high-probability regions under the target model, leading to low probability ratios and low acceptance rates.
b) \textbf{Non-analytic modified distribution.} When a proposed token is rejected, a new token will be drawn from the modified distribution. In continuous distributions, this distribution lacks an analytic expression due to a complex normalizing integral, making direct sampling intractable.

To overcome these challenges, we propose a set of synergistic solutions. \textbf{First}, we address the low acceptance rate via a three-pronged approach: (i) an \textit{approximated acceptance criterion} computed using the probabilities of the samples drawn from the draft model and the target model, respectively, enabling efficient sampling; (ii) a \textit{denoising trajectory alignment} method, grounded in a theory of \textit{reparameterization proximity}, to minimize the distributional divergence while reducing the bias introduced by the approximated acceptance criterion; and (iii) a token pre-filling strategy to stabilize acceptance rates in early generation steps. \textbf{Second}, to sample from the non-analytic modified distribution, we employ an acceptance-rejection sampling scheme~\cite{casella2004generalized}. We derive an appropriate upper bound to avoid computing the complex integral and introduce an easy-to-compute rejection threshold by reusing the property of denoising trajectory alignment, thereby avoiding extra model inference. 

Our continuous speculative decoding can be integrated seamlessly into many existing models, as shown in \Figref{fig:teaser}. 
We validate the effectiveness of our algorithm on various continuous visual AR models~\cite{li2024mar,ren2025xar,wu2025harmon} at two resolutions (256 \& 512) through qualitative and quantitative evaluations. Specifically, we measure wall-time improvements and report image generation quality using different evaluation metrics, including Fréchet Inception Distance (FID)~\cite{heusel2017fid} and Inception Score (IS)~\cite{salimans2016is}. Extensive experiments show that our algorithm achieves over $2\times$ inference speedup while maintaining generation quality.

Our contributions can be summarized as follows:
\begin{itemize}
    \item We are the first to propose continuous speculative decoding, bridging the gap of speculative decoding to continuous distributions and enabling substantial acceleration of continuous visual AR models.
    \item We address the problem of low acceptance rate in continuous speculative decoding via a novel approximated criterion, denoising trajectory alignment, and token pre-filling.
    \item We enable tractable sampling from the modified distribution via a tailored acceptance-rejection sampling scheme with a proper upper bound.
    \item We validate our proposed method into three existing continuous visual AR models without extra training or architectural changes. Extensive experiments show that it achieves over $2\times$ inference speedup while fully maintaining generation quality.
\end{itemize}

%% file: figure_tex/figure_intro.tex
\begin{figure}[t!]
    \centering
    \includegraphics[width=1.0\columnwidth]{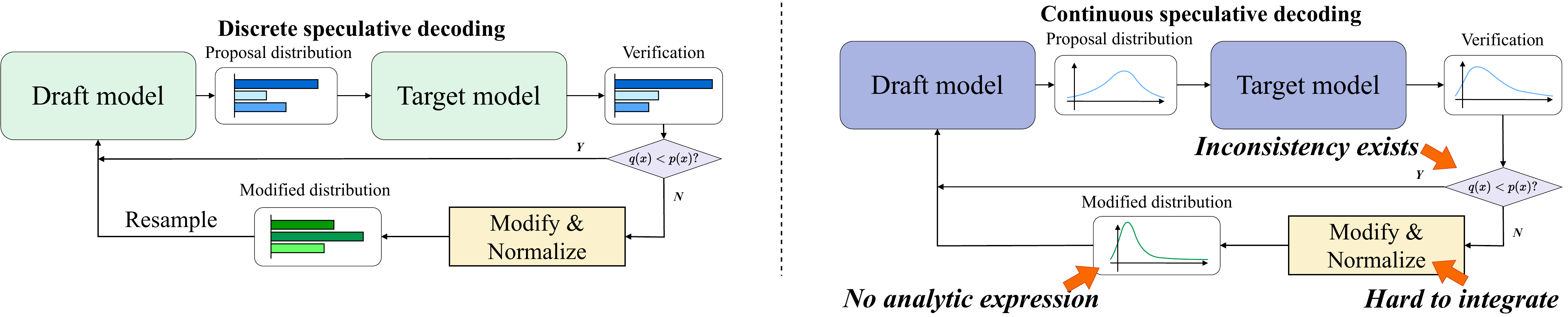}
    \caption{
    Comparison between discrete and continuous speculative decoding. Discrete situation offers the convenience of directly computing probabilities and simply sampling from modified distributions. In contrast, continuous situation faces challenges in the inconsistency of output distributions, leading to low acceptance criterion as well as low acceptance rate, and the modified distributions without analytic expression, caused by complex integral.
    } 
    \label{fig:figure_intro}
\end{figure}

%% file: sec/2_related_work.tex
\section{Related Work}
\label{sec:related}

\subsection{Autoregressive Image Generation}
Autoregressive (AR) models are widely used in image generation. Early works perform generation at pixel level using CNN~\cite{van2016pixelcnn,chen2018pixelsnail}, RNN~\cite{van2016pixelrnn} and Transformer~\cite{parmar2018image,chen2020generative}.
VAR~\cite{VAR} modifies the autoregressive paradigm into next-scale prediction to gradually increase the scale of predictions. Similar to the autoregressive language model, AR image generation through discrete token prediction is scalable to text-conditioned image generation~\cite{liu2024lumina,sun2024llamagen,yu2023magvitv2}. However, training discrete image tokenizer is difficult, and its ability to convey detailed visuals is still questionable~\cite{yu2023magvit,mentzer2023finite}. GIVT~\cite{tschannen2025givt} represents continuous tokens via Gaussian mixture models. MAR~\cite{li2024mar} and DisCo-Diff~\cite{xu2024discodiff} generate tokens via diffusion process~\cite{ho2020ddpm} conditioned by the autoregressive model. HART~\cite{tang2024hart} employs discrete and continuous tokenizer to generate images, with classification for discrete tokens and denoising for the residual between primitive visual tokens and discrete tokens. xAR~\cite{ren2025xar} extends the conception of token and reformulates discrete token classification as continuous entity regression. However, autoregressive models suffer from heavy inference overhead. The inference speed is slowed down by step-by-step generation.

\subsection{Speculative Decoding}
Speculative decoding~\cite{leviathan2023fast,chen2023acc} achieves lossless acceleration by verifying the draft model with the target model. Following this, previous works mainly focus on reducing draft model overhead and strengthening the consistency between the draft and target models.
SpecInfer~\cite{miao2023specinfer} employs multiple small draft models and aggregates their predictions into a tree structure to be verified through tree-based parallel decoding.
Eagle~\cite{li2024eagle,li2024eagle2} improves the draft accuracy through the prediction at the feature level instead of the token level to tackle the feature uncertainty problem.
Jacobi iteration is also employed to reduce inference overhead in the decoding process~\cite{santilli2023pardec,zhao2024lookahead,zhao2024ouroboros,kou2024cllms}.
Online Speculative Decoding~\cite{liu2023online} and DistillSpec~\cite{zhou2023distillspec} align the output from the draft model with the target model with more training. 
Speculative decoding can achieve lossless acceleration theoretically, but the generation quality may be affected under a larger speed-up ratio. BiLD~\cite{kim2024bild} proposes a more relaxed acceptance condition. Besides, finetuning the target model can also improve generation quality~\cite{cai2024medusa,kou2024cllms,yi2024space}.

Current literature explores applying speculative decoding to accelerate autoregressive image generation. Methods like SJD~\cite{teng2024sjd} combine speculative verification with Jacobi iterations to boost speed without sacrificing variety. Furthermore, LANTERN~\cite{jang2024lantern} and LANTERN++~\cite{park2025lanternpp} leverage relaxed candidate selection to manage distribution ambiguity and preserve image quality. Despite reducing total inference steps, these techniques are fundamentally limited to discrete token spaces. This paper addresses this limitation by proposing a continuous speculative decoding framework specifically designed for continuous autoregressive models.

%% file: figure_tex/figure_main.tex
\begin{figure}[t!]
    \centering
    \includegraphics[width=1.0\columnwidth]{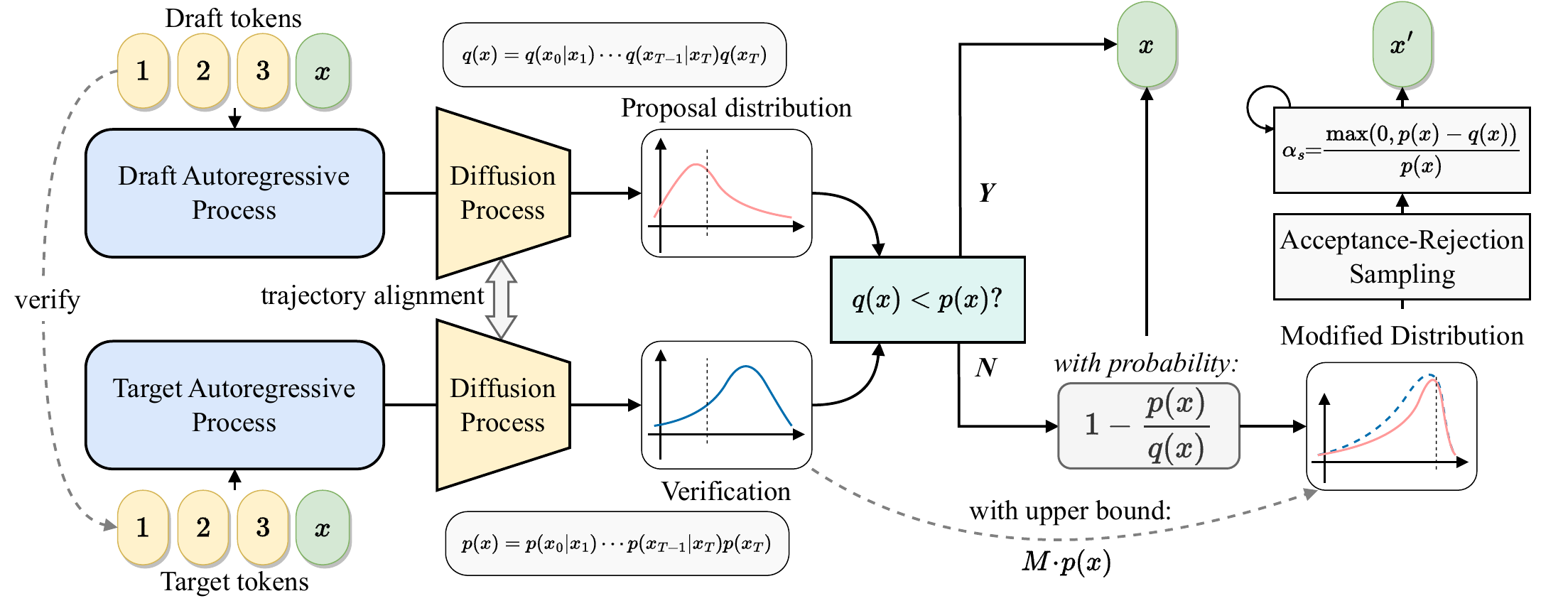}
    \caption{
    The overview of continuous speculative decoding. The diffusion model component of continuous AR models is leveraged. Tokens $1\sim 3$ are prefix tokens, and token $x$ is to be verified. In this method, $x$ is accepted if $q(x)<p(x)$. Otherwise, $x$ is rejected with probability $1-{p(x)}/{q(x)}$, followed by sampling $x^\prime$ from the modified distribution via acceptance-rejection sampling.
    } 
    \label{fig:figure_task}
\end{figure}

%% file: sec/3_methodology.tex
\section{Methodology}
\label{sec:method}

This section introduces our continuous speculative decoding framework for visual autoregressive models. We first extend discrete speculative decoding to continuous diffusion trajectories, then analyze why direct application leads to low acceptance rates. To mitigate this, we propose denoising trajectory alignment and target-token pre-filling to reduce draft-target inconsistency. Finally, we derive an efficient acceptance--rejection sampling strategy for resampling from the modified distribution.

\subsection{From Discrete to Continuous Speculative Decoding}

We first introduce the discrete form of speculative decoding. It utilizes a draft model $M_q$ with output distribution $q(x)$ and a target model $M_p$ with $p(x)$.
$M_q$ generates a sequence of draft tokens $x_{i:j}=\{x_i,\dots,x_j\}$ where $x\sim q(x)$, which are then verified by $M_p$ in parallel. $x$ is accepted if the acceptance criterion ${p(x)}/{q(x)}>1$; otherwise, it is rejected with probability $1-{p(x)}/{q(x)}$ and resampled from a modified distribution $p^\prime (x)=norm(max(0,p(x)-q(x)))=\frac{max(0,p(x)-q(x))}{\sum_{x^\prime} max(0,p({x^\prime})-q({x^\prime}))}$.

Then, we discuss continuous speculative decoding (see \Figref{fig:figure_task}). In continuous visual AR models~\cite{li2024mar}, the output distribution is typically modeled via a diffusion model~\cite{ho2020ddpm,nichol202ddim}, namely:
\begin{align}
    p(x_{N,0:T},|x_{1:N-1})
    =p(x_{N,T})\prod_{t=1}^Tp(x_{N,t-1}|x_{N,t},x_{1:N-1}),
    \label{eqn:first_define}
\end{align}
where $N$ represents $N$-th autoregressive step. $t\in[0,T]$ is the diffusion timestep. $x_{N,t}$ represents the state of $x$ at autoregressive step $N$ and diffusion timestep $t$. $x_{1:N-1}=\{x_1,x_2,\dots,x_{N-1}\}$ denotes the variables before step $N$. For simplicity, we omit $N$ and $x_{1:N-1}$ and let $Y=x_{0:T}$ to obtain $p(Y)=p(x_T)\prod_{t=1}^Tp(x_{t-1}|x_t)$. 

\textbf{Acceptance Criterion.} The acceptance criterion is defined as the ratio of the probability under target distribution to the one under draft distribution, that is, $p(x_{0:T})/q(x_{0:T})$.
In discrete form, the probability can be directly obtained. But in continuous form, the probability is usually obtained via diffusion process~\cite{ho2020ddpm,song2020denoisingddim}. Specifically, $x_{0:T}$ is sampled from draft model via reverse diffusion (denoising) process. So the ratio is given by:
\begin{equation}
    \frac{p(Y)}{q(Y)}=\frac{p(x_T)\prod_{t=1}^T p(x_{t-1}|x_t)}{q(x_T)\prod_{t=1}^T q(x_{t-1}|x_t)},
    \label{eqn:raw_ratio}
\end{equation}
where $x_T$ is sampled from Gaussian noise and $p(x_{t-1}|x_t)$ is approximated as a Gaussian distribution through a neural network with parameter $\theta$~\cite{nichol202ddim}, and $\mu_\theta(x_t,t)$ and $\Sigma_\theta(x_t,t)$ are mean and variance predicted by $\theta$, that is:
\begin{equation}
    p(x_{t-1}|x_t)=\mathcal{N}(x_{t-1};\mu_\theta(x_t,t),\Sigma_\theta(x_t,t)).
    \label{eqn:normal}
\end{equation}

However, this approach is impractical because the acceptance rate is low. As shown in Table~\ref{tab:likelihood_ratio} and discussed in Appendix~\ref{sec:approximation}, the draft model’s trajectory $x_{0:T}$ inherently diverges from the target model’s expected trajectory. In each denoising step, samples drawn from the draft model's distribution $q$ are unlikely to fall near $\mu$ of the target distribution $p$, which results in a low single-step ratio $p/q$. As the multi-step denoising process proceeds, the overall $p(Y)/q(Y)$ becomes extremely small.

Therefore, our work adopts a more practical approximation: the ratio of the joint probabilities $p(Y_p)/q(Y_q)$, where both $Y_p$ and $Y_q$ share the same $x_0$. This ratio serves as a surrogate for the shared path ratio. Then, the ratio is calculated through:
\begin{equation}
    \frac{p(Y_p)}{q(Y_q)}\coloneqq\frac{p(x_T^p)\prod_{t=1}^T p(x_{t-1}^p|x_t^p)}{q(x_T^q)\prod_{t=1}^T q(x_{t-1}^q|x_t^q)},\quad x_0^p=x_0^q=x_0.
    \label{eqn:ratio}
\end{equation}

\input{table/maginal_distribution}

\textbf{Modified Distribution.} When a draft token is rejected, a new token is resampled from the modified distribution $p^\prime (x)$. Following the discrete formulation, $p^\prime (x)$ is derived from the normalization of $max(0,p(x)-q(x))$. Therefore, replacing the summation in the normalization denominator of the discrete form with integral yields the continuous form, namely:
\begin{equation}
    p^\prime (Y)=\frac{max(0,p(Y)-q(Y))}{Z},\,\text{where}\, Z=\int_{Y^\prime}max(0,p(Y^\prime)-q(Y^\prime))dY^\prime.\label{eqn:modified}
\end{equation}

However, in practical operation, an extremely low acceptance rate emerges, as shown in Table~\ref{tab:acceptance}, thereby leading to poor acceleration performance. To address this, we propose the following methods.

\subsection{Mitigating distribution inconsistency}

Under the acceptance criterion given by \Eqref{eqn:ratio}, the acceptance rate $\alpha$ is extremely small (nearly 0\%). We attribute this to two sources of inconsistency in continuous visual AR models: \textbf{1) inconsistency in diffusion process}, and \textbf{2) inconsistency in autoregressive process}.

\textbf{Inconsistency in Diffusion Process.} Significant inconsistency exists in the denoising process. As illustrated in \Figref{fig:figure_alignment}, using the approximated acceptance criterion, draft and target denoising trajectories diverge to different outputs. The output distance between the two distributions is large. Besides, the approximated acceptance criterion is also biased relative to the original probability ratio, and therefore it does not preserve the unbiasedness guarantee of the original speculative decoding.

\input{figure_tex/denoise_align_and_prefill}
We propose denoising trajectory alignment to enhance the consistency of the output distributions and reduce the degree of bias. Note that in \Eqref{eqn:normal}, $x_{t-1}^p$ is obtained via reparameterization given by $x_{t-1}^p=\sqrt{\Sigma^p_\theta(x^p_t,t)}\cdot\varepsilon_t^p+\mu^p_\theta(x^p_t,t)$, where $\varepsilon_t^p\sim\mathcal{N}(0,\text{I})$ (same for $x_{t-1}^q$). Denoising trajectory alignment can reduce the expected distance between $x^p_{t-1}$ and $x^q_{t-1}$, promised by the following theorem.

\begin{restatable}[Reparameterization Proximity]{thm}{proximity}\label{thm:proximity}
Setting $\varepsilon^p_t=\varepsilon^q_t$ in reparameterization reduces the expected distance $\mathbb{E}\left[\left\|x_{t-1}^q - x_{t-1}^p\right\|^2\right]$ between $x^p_{t-1}$ and $x^q_{t-1}$ by $2\cdot\text{tr}\left[\sqrt{\Sigma^q_t\Sigma^p_t}\right]$.
\end{restatable}

Detailed proofs can be found in Appendix~\ref{sec:proof}. Note that:
\begin{align}
    p(x_{t-1}|x_t)&=\frac{\exp{\{\frac{1}{2}[x_{t-1}\!\!-\!\!\mu_\theta(x_t,t)]^T\Sigma^{-1}_\theta(x_t,t)[x_{t-1}\!-\!\mu_\theta(x_t,t)]\}}}{(\sqrt{2\pi})^n\sqrt{|\Sigma_\theta(x_t,t)|}}\notag\\
    &=\frac{1}{(\sqrt{2\pi})^n\sqrt{|\Sigma_\theta(x_t,t)|}}\exp{\{\frac{1}{2}{\varepsilon_t}^T\varepsilon_t\}},\quad x_t\in\{x^p_t,x^q_t\}, \varepsilon_t\in\{\varepsilon^p_t,\varepsilon^q_t\}.
\end{align}
The ratio ${p(x^p_{t-1}|x^p_t)}/{q(x^q_{t-1}|x^q_t)}$ given $\varepsilon
_t=\varepsilon^p_t=\varepsilon^q_t$ is:
\begin{align}
    \frac{p(x^p_{t-1}|x^p_t)}{q(x^q_{t-1}|x^q_t)}
    &=\frac{\frac{1}{(\sqrt{2\pi})^n\sqrt{|\Sigma^p_t|}}\exp{\{\frac{1}{2}\varepsilon_t^T\varepsilon_t\}}}{\frac{1}{(\sqrt{2\pi})^n\sqrt{|\Sigma^q_t|}}\exp{\{\frac{1}{2}\varepsilon_t^T\varepsilon_t\}}}
    =\frac{\sqrt{|\Sigma^q_t|}}{\sqrt{|\Sigma^p_t|}}.
\end{align}
For simplicity, we define:
\begin{equation}
    \Sigma=\prod_{t=2}^T\sqrt{|\Sigma^q_{t}|}/\prod_{t=2}^T\sqrt{|\Sigma^p_{t}|}.
    \label{eqn:term}
\end{equation}
Substituting $\Sigma$ into \Eqref{eqn:ratio} makes (assuming $p(x_T^p)=q(x_T^q)$):
\begin{align}
    \frac{p(Y_p)}{q(Y_q)}
    =\frac{p(x_T^p)p(x_0|x_1^p)\prod_{t=2}^Tp(x^p_{t-1}|x^p_t)}{q(x_T^q)q(x_0|x_1^q)\prod_{t=2}^Tq(x^q_{t-1}|x^q_t)}
    =\frac{p_\theta(x_0|x^p_1)}{q_\theta(x_0|x^q_1)}\cdot \Sigma.
        \label{eqn:eliminate}
\end{align}

Therefore, at each step, employing the same $\varepsilon_t$ reduces the expected distance by $2\cdot\text{tr}\left[\Sigma^q_t\Sigma^p_t\right]$, which enables the two models to generate closer samples and finally increases the acceptance rate, as shown in Table~\ref{tab:likelihood_ratio}.

\textbf{Inconsistency in Autoregressive Process.} Inconsistency also arises in autoregressive steps. \Figref{fig:figure_prefilling} shows that acceptance rate $\alpha$ is very low (5\%) at the initial AR steps, and it increases progressively as AR steps grow. This stems from the different draft and target prefix embeddings~\cite{li2024mar}. Owing to this, the AR models naturally yield divergent predictions, which in turn leads to low $\alpha$. As the AR steps increase, the inputs of the two models gradually converge, thereby improving consistency between their outputs and finally raising the $\alpha$.

To address this, we propose pre-filling a portion (e.g., $5\%$) of tokens from the target model to ensure a consistent prefix. This does not increase inference latency, as speculative decoding at a low acceptance rate is functionally equivalent to the target model step-by-step decoding~\cite{leviathan2023fast}. Furthermore, pre-filling improves the overall acceptance rate.

Finally, $p(Y_p)/q(Y_q)$ can be computed with the help of denoising trajectory alignment and token pre-filling to obtain a considerable acceptance rate.

\subsection{Resample from the modified distribution}\label{sec:modify}
The simplified illustration of \Eqref{eqn:modified} is shown in \Figref{fig:figure_modified}. Unlike discrete form, where $\sum max(0,p(x)-q(x))$ can be directly computed, the analytic expression of $Z$ can't be computed because it involves the integral of the product of a series of Gaussian distributions given by \Eqref{eqn:ratio} and \ref{eqn:normal}. Therefore, $p^\prime (Y)$ cannot be directly sampled.

To tackle this problem, a viable approach is acceptance–rejection sampling~\cite{casella2004generalized}, which first samples from a proposal distribution, and then calculates a pre-defined rejection threshold $\alpha_s$ and samples $\epsilon\sim U(0,1)$. If $\epsilon<\alpha_s$, the sample is accepted, otherwise it is rejected and sampling from the proposal distribution will be repeated until the sample is accepted.

To realize this sampling method, we first define $\alpha_s$ as:
\begin{equation}
    \alpha_s=\frac{p^\prime(Y)}{M\cdot p(Y)},
    \label{eqn:alpha}
\end{equation}
where $p(Y)$ is the target distribution, and $M$ is the upper bound factor that holds $M\cdot p(Y)\geq p^\prime(Y)$ for any $Y$. Given $max(0,p(Y)-q(Y))\le p(Y)$, $M$ is set to $1/Z$ considering that:
\begin{equation}
    p^\prime(Y)=\frac{max(0,p(Y)-q(Y))}{Z}\leq\frac{p(Y)}{Z}\mapsto M\cdot p(Y).
\end{equation}
We substitute $M=1/Z$ into \Eqref{eqn:alpha} to eliminate $Z$:
\begin{equation}
    \alpha_s=\frac{max(0,p(Y)-q(Y))/Z}{p(Y)/Z}=\frac{max(0,p(Y)-q(Y))}{p(Y)}.
    \label{eqn:rejsample}
\end{equation}
\Eqref{eqn:rejsample} gives the analytic expression of $\alpha_s$ without calculating $Z$. However, in its naive implementation, repetitive diffusion model inference is needed to sample $p(Y)$. It brings heavy extra overhead. To tackle this problem, denoising trajectory alignment is introduced to simplify \Eqref{eqn:rejsample}. Accordingly, we have derived the following corollary.

\begin{restatable}[Easy-to-Compute Rejection Threshold]{cor}{simplified}\label{thm:simplified}
With denoising trajectory alignment introduced from \Eqref{eqn:eliminate}, the rejection threshold $\alpha_s$ has the easy-to-compute form:
\begin{equation}
    \alpha_s=\frac{max(0,\Sigma\cdot p_\theta(x_0|x^p_1)-q_\theta(x_0|x^q_1))}{\Sigma\cdot p_\theta(x_0|x^p_1)}.
\end{equation}
\end{restatable}

See Appendix~\ref{sec:proof} for detailed proofs. 
In this form, $x_0$ is sampled from $p_\theta(x_0|x^p_1)$, which is merely a Gaussian distribution defined by \Eqref{eqn:normal}, thus avoiding extra model inference. We can compute $\alpha_s$ by gathering $\Sigma$ and sampling $x_0$ from the Gaussian distribution, thereby completing the acceptance-rejection sampling to obtain samples equivalent to those derived from sampling $p^\prime(Y)$.

%% file: table/maginal_distribution.tex
\begin{table}[t]
    \centering
    \caption{Probability ratio of different calculation approaches. $Y$ denotes the shared denoising trajectory of the sample. $Y_p$ and $Y_q$ represent the samples generated by the target model and the draft model through the denoising process, respectively.} 
    \label{tab:likelihood_ratio} 
    \begin{tabular}{lcc}
        \toprule
        \textbf{Probability ratio} & \textbf{Value} & \textbf{Acceptance rate} \\
        \midrule
        $p(Y)/q(Y)$ & $5.33 \times 10^{-23}$ & $0.0\%$ \\
        $p(Y_p)/q(Y_q)$, w/o align & $0.067$ & $14\%$ \\
        $p(Y_p)/q(Y_q)$, w/ align & $1.86$ & $32\%$ \\
        \bottomrule
    \end{tabular}
    
    \label{tab:marginal}
\end{table}

%% file: figure_tex/denoise_align_and_prefill.tex
\begin{figure}[t] 
    \centering 

    \begin{minipage}[t]{0.62\textwidth} 
        \centering 
        \includegraphics[width=\textwidth]{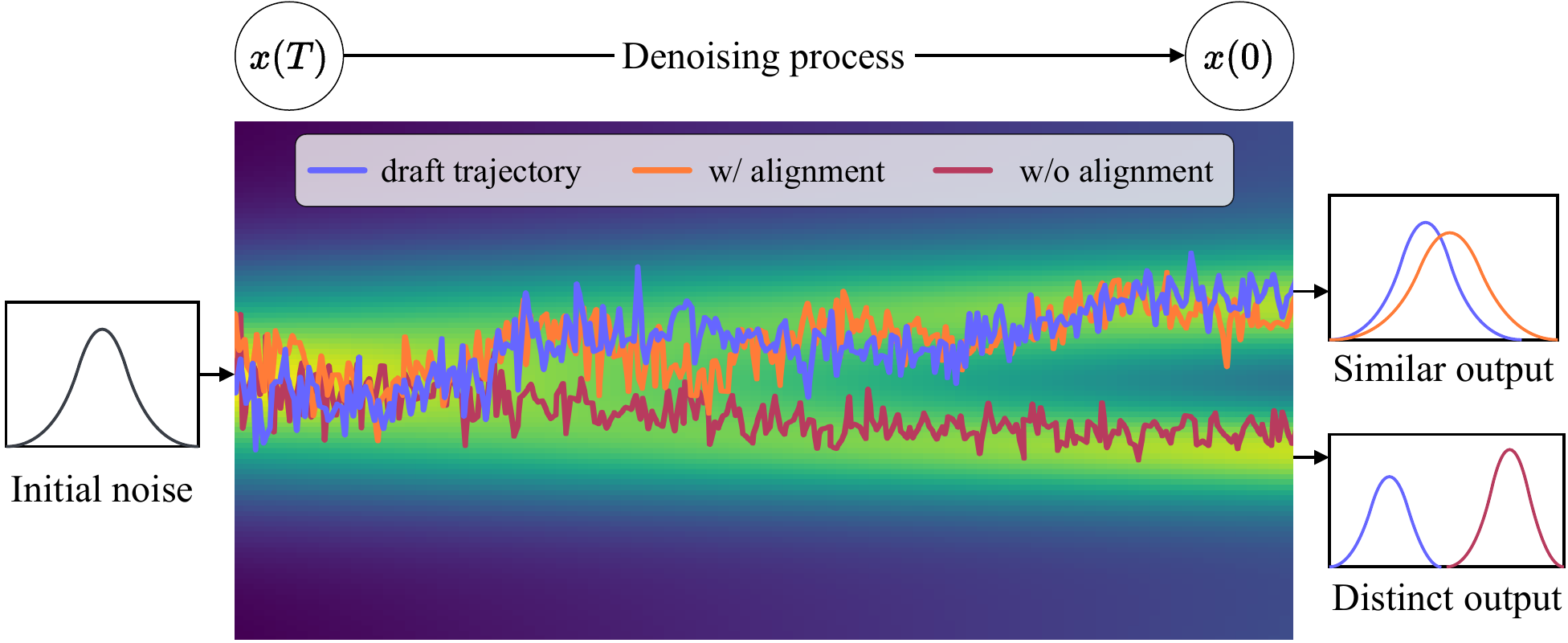}
        \caption{
    Illustration of denoising trajectory alignment. The denoising process maps the noise distribution to data distribution through gradual denoising. These denoising steps form a trajectory. The aligned trajectory (orange curve) leads to a similar output distribution, while the unaligned one (red curve) produces a far-away one, obtaining low ${p(Y_p)}/{q(Y_q)}$.
    } 
    \label{fig:figure_alignment}
    \end{minipage}
    \hfill
    \begin{minipage}[t]{0.36\textwidth}
        \centering
        \includegraphics[width=\textwidth]{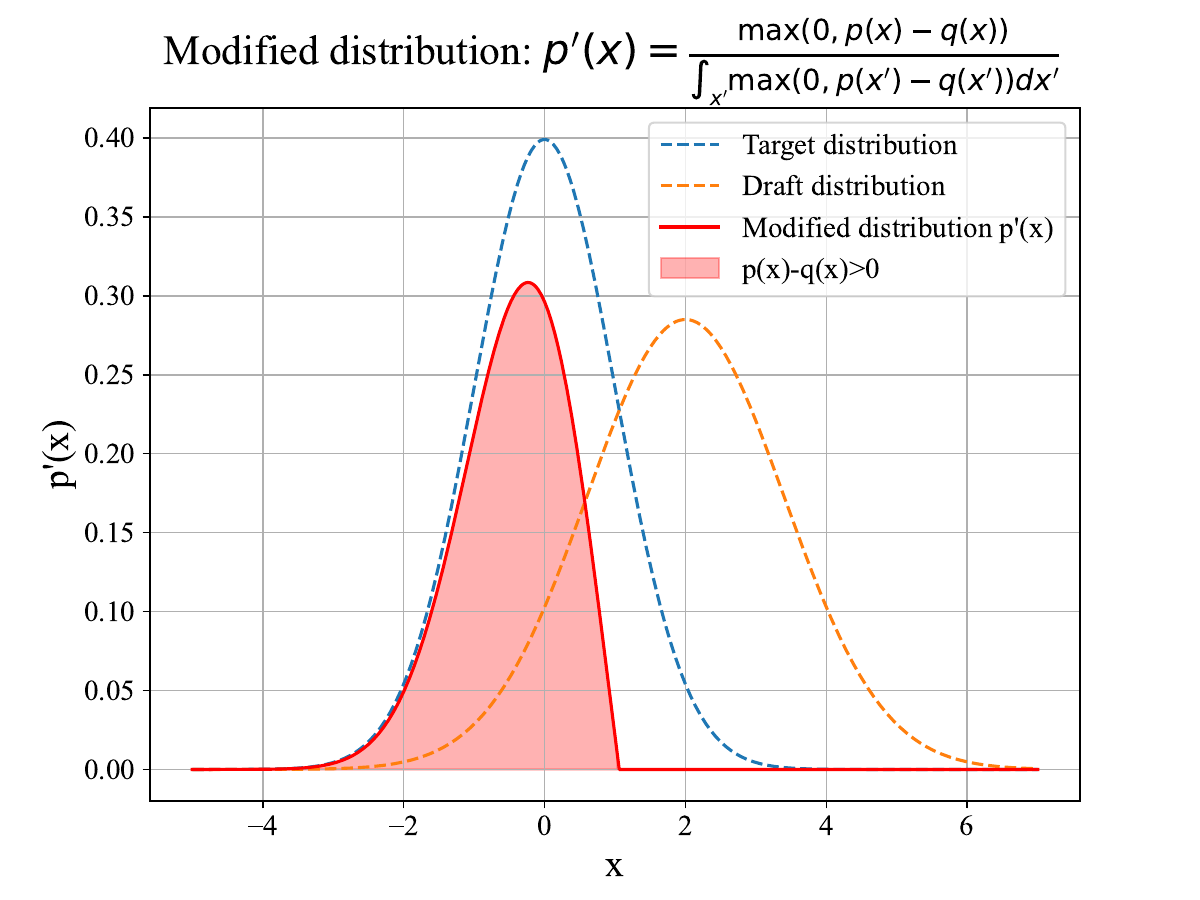}
        \caption{
    Illustration of distributions. Dashed lines: draft and target distributions. Red area: modified distribution (unnormalized, omitting $Z$ for simplicity), whose integral lack an analytic expression.
    } 
    \label{fig:figure_modified}
    \end{minipage}
\end{figure}

%% file: sec/4_experiment.tex
\section{Experiment}
\label{sec:experiment}

\input{figure_tex/figure_vis}
\input{table/speed_up}
\subsection{Implementation Details}
We evaluate our method across several open-source continuous visual autoregressive models: MAR~\cite{li2024mar}, xAR~\cite{ren2025xar}, and Harmon~\cite{wu2025harmon}. Experiments are conducted on ImageNet~\cite{deng2009imagenet} for MAR and xAR at $256\times 256$ resolution, while Harmon is tested at both $256\times 256$ and $512\times 512$ resolutions, leveraging its diverse pre-training data. We pair draft models—MAR-B (208M), xAR-B (172M), and Harmon-0.5B—with their corresponding target models: MAR-H (943M), xAR-H (1.1B), and Harmon-1.5B. Performance is reported using Fréchet Inception Distance (FID)~\cite{heusel2017fid} and Inception Score (IS)~\cite{salimans2016is} for MAR and xAR, and FID, CLIPScore~\cite{hessel2021clipscore}, and Geneval~\cite{ghosh2023geneval} for Harmon. Speedups are reported based on wall-clock time on a single NVIDIA A100 GPU. More details of experiment settings, ablation studies and quantitative results can be found in Appendix.
\input{table/fid_and_is}

\subsection{Main Results}
\paragraph{Speedup results.}
As shown in Tables~\ref{tab:speed_up} and~\ref{tab:harmon_combined_speedup}, the speedup ratio and acceptance rate exhibit a positive correlation with batch size, with draft lengths ranging from $8$ to $32$. This trend highlights the growing efficacy of speculative decoding in higher-throughput scenarios. Specifically, our algorithm achieves substantial speedups of $2.33\times$ on MAR, $2.72\times$ on xAR, and $2.54\times$ on Harmon, demonstrating robust performance across diverse model architectures.

\paragraph{Quantitative results.}
Tables~\ref{tab:exp_fid_is} and~\ref{tab:xar_eval} present the class-conditioned and unconditioned FID and IS metrics for our continuous speculative decoding applied to MAR and xAR. For Harmon, FID on MSCOCO~\cite{lin2014microsoftcoco} and MJHQ~\cite{li2024playgroundmjhq}, alongside CLIPScore~\cite{hessel2021clipscore} and Geneval~\cite{ghosh2023geneval} are reported in Tables~\ref{tab:harmon_combined_eval} and ~\ref{tab:geneval_eval}. We conduct multiple experiments and report the average performance and the standard deviation to ensure the reliability of our conclusions. These results demonstrate that our algorithm significantly preserves the quality of generated images, which will be further discussed in subsequent sections. Thus, our approach offers a robust solution for efficient and reliable model inference.

\paragraph{Qualitative results.}
Visual comparisons in Figures~\ref{fig:figure_vis} and~\ref{fig:figure_comparasion} demonstrate the superior image quality produced by our algorithm. While Figure~\ref{fig:figure_vis} highlights the qualitative results of our method, Figure~\ref{fig:figure_comparasion} provides a direct contrast with the baseline MAR-H (autoregressive steps=$256$) under varying draft lengths $\gamma$. These visualizations confirm that our approach achieves significant acceleration while preserving image fidelity.

\input{table/acceptance_and_prefill}
\input{figure_tex/draft_vis_and_curve}
\input{figure_tex/figure_prefilling}
\subsection{Ablation Study}

\paragraph{The $\alpha$ vs. $\gamma$.}
The relationship between acceptance rate $\alpha$ and draft length $\gamma$ on MAR model is depicted in \Figref{fig:figure_acceptance_gamma}. As the length of the draft increases, the acceptance rate tends to decline. This observation suggests that while longer drafts can substantially mitigate inference overhead, they are intrinsically constrained by the capabilities of the draft model itself. Consequently, an increase in the number of draft lengths is associated with greater deviations from the target model's distribution, ultimately leading to reduced acceptance rates.

\paragraph{Effectiveness of denoising trajectory alignment.}
Table~\ref{tab:acceptance} shows the acceptance rate and the average distance of each draft and target token with and without denoising trajectory alignment. The results show that, denoising trajectory alignment reduces the distance between the draft tokens and the target tokens. Before alignment, the distance between them reaches $>2$, leading to quite small $\alpha$. As the distance is reduced to $\sim1$, the probability ratio increases, further increasing the $\alpha$ to $>30\%$.

\Figref{fig:figure_align_vis} illustrates the image generated with and without alignment. Without denoising trajectory alignment, using trajectories sampled from the draft and target models independently leads to significant divergence, which not only reduces the token acceptance rate but also causes deformations and artifacts. This quality degradation is attributable to the fact that the approximate acceptance criterion is biased relative to the original probability ratio. With the application of denoising trajectory alignment, these negative effects are alleviated: divergence is significantly reduced, token acceptance rates are improved, and deformations and artifacts are effectively eliminated.

\input{figure_tex/figure_align_vis_and_prefill_vis}

\paragraph{Influence of pre-filled tokens.}
The ablation study of pre-filling ratios at \(0\%\), \(5\%\), and \(15\%\) on MAR model is illustrated in \Figref{fig:figure_prefilling}. Pre-filling can compensate for the low acceptance rates observed during the initial stages of autoregressive sampling and enhance the overall acceptance rate, as shown in Table~\ref{tab:prefilling}. Moreover, \Figref{fig:figure_prefil_vis} shows the visualizations under
different pre-filling ratios. Notably, the discrepancies between the draft and the target model result in certain artifacts and reduced image quality at 0\% pre-filling. However,
introducing a modest proportion of pre-filled tokens from
the target model has effectively mitigated these artifacts. As
the pre-filling ratio increases, the advantages conferred by
this approach exhibit diminishing returns.

%% file: figure_tex/figure_vis.tex
\begin{figure*}[t!]
    \centering
    \includegraphics[width=1.0\linewidth]{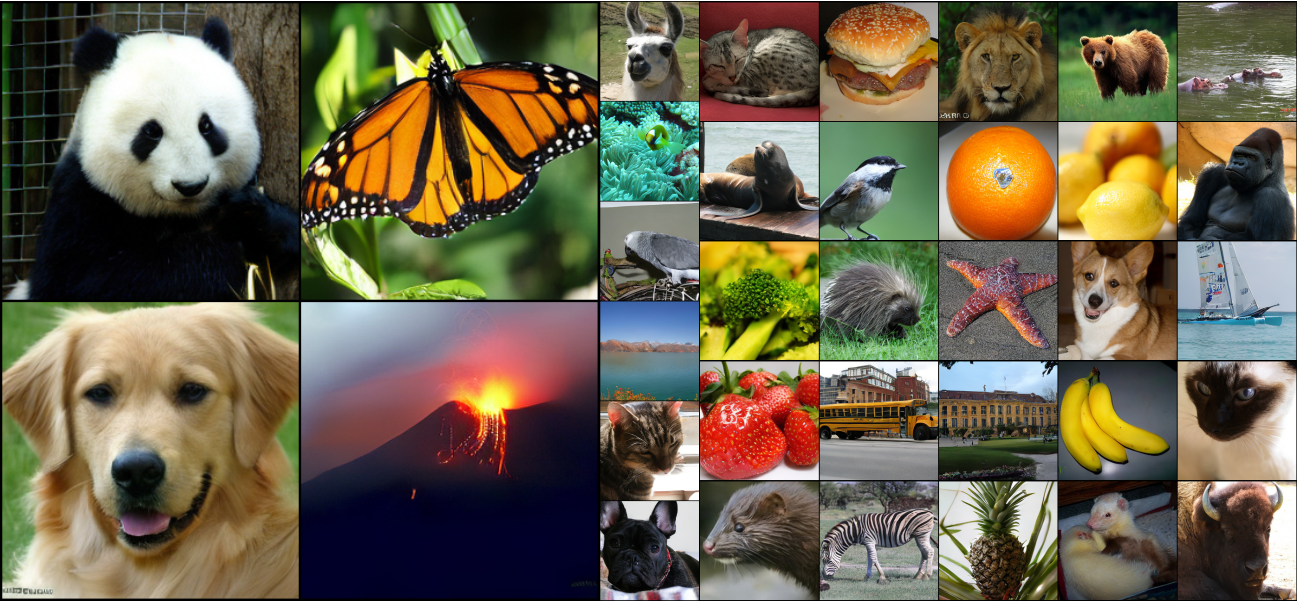}
    \caption{
    Images generated by MAR with continuous speculative decoding.
    } 
    \label{fig:figure_vis}
\end{figure*}

%% file: table/speed_up.tex
\begin{table}[t!]
    \small
    \centering
    \caption{Results of speedup ratio and acceptance rate $\alpha$ on MAR and xAR under different draft lengths and batch sizes. The bs refers to batch size.}
    \setlength{\tabcolsep}{5.8pt} 
    \renewcommand{\arraystretch}{1.1} 
    \begin{tabular}{cccc|cccc} 
        \toprule 
        \multirow{2}*{$M_p$} & \multirow{2}*{$M_q$} & \multirow{2}*{$\gamma$} & \multirow{2}*{$\alpha$} & \multicolumn{4}{c}{Speedup ratio}\\\cmidrule(lr){5-8} &     & &                 
        &bs=1&bs=8&bs=128&bs=256\\
        \midrule 
        MAR-\scriptsize H & MAR-\scriptsize B & 32 & 0.19 & \textbf{1.44$\times$} & \textbf{1.61$\times$} & \textbf{2.17$\times$} & \textbf{2.33$\times$} \\
        MAR-\scriptsize H & MAR-\scriptsize B & 16 & 0.26 & 1.37$\times$ & 1.51$\times$ & 2.07$\times$ & 2.20$\times$ \\
        MAR-\scriptsize H & MAR-\scriptsize B & 8 & 0.27 & 1.26$\times$ & 1.44$\times$ & 1.88$\times$ & 1.96$\times$ \\
        MAR-\scriptsize H & MAR-\scriptsize B & 4 & 0.30 & 1.11$\times$ & 1.20$\times$ & 1.56$\times$ & 1.62$\times$ \\
        \midrule
        xAR-\scriptsize H & xAR-\scriptsize B & 32 & 0.22 & \textbf{1.77$\times$} & \textbf{2.10$\times$} & \textbf{2.52$\times$} & \textbf{2.72$\times$}  \\
        xAR-\scriptsize H & xAR-\scriptsize B & 16 & 0.26 & 1.58$\times$ & 2.06$\times$ & 2.31$\times$ & 2.61$\times$  \\
        xAR-\scriptsize H & xAR-\scriptsize B & 8 & 0.29 & 1.61$\times$ & 1.86$\times$ & 2.07$\times$ & 2.18$\times$  \\
        xAR-\scriptsize H & xAR-\scriptsize B & 4 & 0.36 & 1.30$\times$ & 1.62$\times$ & 1.92$\times$ & 2.11$\times$  \\
        \bottomrule 
    \end{tabular}
    
    \label{tab:speed_up}
\end{table}

\begin{table}[t]
    \small 
    \centering 
    \caption{Results of speedup ratio and acceptance rate $\alpha$ on Harmon under different resolutions, draft lengths, and batch sizes (bs). Due to CUDA out-of-memory, this set of experiments opt for smaller batch sizes.} 
    \setlength{\tabcolsep}{4.2pt} 
    \renewcommand{\arraystretch}{1.2} 
    \begin{tabular}{c|cccc|cccc} 
        \toprule
        \multirow{2}*{Resolution} & \multirow{2}*{$M_p$} & \multirow{2}*{$M_q$} & \multirow{2}*{$\gamma$} & \multirow{2}*{$\alpha$} & \multicolumn{4}{c}{Speedup ratio}\\\cmidrule(lr){6-9} 
        & & & & & bs=1 & bs=8 & bs=16 & bs=32 \\
        \midrule
        \multirow{4}*{256} & Harmon-\scriptsize H & Harmon-\scriptsize B & 32 & 0.17 & \textbf{1.47$\times$} & \textbf{1.67$\times$} & \textbf{1.88$\times$} & \textbf{2.05$\times$} \\
        & Harmon-\scriptsize H & Harmon-\scriptsize B & 16 & 0.21 & 1.41$\times$ & 1.58$\times$ & 1.78$\times$ & 1.93$\times$ \\
        & Harmon-\scriptsize H & Harmon-\scriptsize B & 8 & 0.25 & 1.29$\times$ & 1.44$\times$ & 1.60$\times$ & 1.72$\times$ \\
        & Harmon-\scriptsize H & Harmon-\scriptsize B & 4 & 0.33 & 1.11$\times$ & 1.22$\times$ & 1.33$\times$ & 1.42$\times$ \\
        \midrule 
        \multirow{4}*{512} & Harmon-\scriptsize H & Harmon-\scriptsize B & 32 & 0.15 & \textbf{1.63$\times$} & \textbf{1.94$\times$} & \textbf{2.23$\times$} & \textbf{2.54$\times$} \\
        & Harmon-\scriptsize H & Harmon-\scriptsize B & 16 & 0.23 & 1.55$\times$ & 1.83$\times$ & 2.09$\times$ & 2.35$\times$ \\
        & Harmon-\scriptsize H & Harmon-\scriptsize B & 8 & 0.25 & 1.41$\times$ & 1.65$\times$ & 1.85$\times$ & 2.05$\times$ \\
        & Harmon-\scriptsize H & Harmon-\scriptsize B & 4 & 0.38 & 1.20$\times$ & 1.37$\times$ & 1.50$\times$ & 1.63$\times$ \\
        \bottomrule
    \end{tabular}
    \label{tab:harmon_combined_speedup} 
    \vspace{0.4cm}
\end{table}







%% file: table/fid_and_is.tex
\begin{table}[t!]

    \small

    \centering

    \caption{Evaluation of FID and IS on unconditional and conditional generation, compared with original MAR-L and MAR-H models. Our method achieves acceleration while maintaining performance within a reasonable interval.}

    \setlength{\tabcolsep}{6pt} 

    \renewcommand{\arraystretch}{1.0} 


    \begin{tabular}{cc|cc|cc} 

        \toprule 

        \multirow{2}*{$M_p$} & \multirow{2}*{$M_q$} & \multicolumn{2}{c|}{w/o CFG} & \multicolumn{2}{c}{w/ CFG} \\

        \cmidrule(lr){3-4} \cmidrule(lr){5-6}

                              &                      & FID$\downarrow$ & IS$\uparrow$                    & FID$\downarrow$ & IS$\uparrow$ \\

        \midrule 

        \multicolumn{2}{c|}{MAR-\scriptsize L}                   & 2.60 & 221.4 & 1.78 & 296.0 \\

        MAR-\scriptsize L & MAR-\scriptsize B & 2.59\scriptsize{$\pm$0.04} & 218.4\scriptsize{$\pm$3.4} & 1.81\scriptsize{$\pm$0.05} & 303.7\scriptsize{$\pm$4.3} \\

        \midrule 

        \multicolumn{2}{c|}{MAR-\scriptsize H}                    & 2.35 & 227.8 & 1.55 & 303.7 \\

        MAR-\scriptsize H & MAR-\scriptsize B & 2.36\scriptsize{$\pm$0.05} & 228.5\scriptsize{$\pm$2.2} & 1.60\scriptsize{$\pm$0.05} & 301.6\scriptsize{$\pm$2.6} \\

        MAR-\scriptsize H & MAR-\scriptsize L & 2.34\scriptsize{$\pm$0.04} & 228.9\scriptsize{$\pm$2.8} & 1.57\scriptsize{$\pm$0.04} & 301.4\scriptsize{$\pm$2.5} \\

        \bottomrule 

    \end{tabular}

    \label{tab:exp_fid_is}

    \vspace{0.4cm}

\end{table}

\begin{table}[t!]

    \small

    \centering

    \caption{Evaluation of FID and IS for xAR models. The results show that using different draft models maintains generation quality (FID and IS) within a stable range compared to the target models alone.}

    \setlength{\tabcolsep}{10pt} 

    \begin{tabular}{cc|cc} 

        \toprule 

        $M_p$ & $M_q$ & FID$\downarrow$ & IS$\uparrow$ \\

        \midrule 

        \multicolumn{2}{c|}{xAR-L} & 1.87 & 274.8 \\

        xAR-L & xAR-B & 1.88\scriptsize{$\pm$0.08} & 270.4\scriptsize{$\pm$6.9} \\

        \midrule 

        \multicolumn{2}{c|}{xAR-H} & 1.79 & 288.9 \\

        xAR-H & xAR-B & 1.82\scriptsize{$\pm$0.07} & 286.1\scriptsize{$\pm$4.2} \\

        xAR-H & xAR-L & 1.78\scriptsize{$\pm$0.07} & 287.7\scriptsize{$\pm$2.7} \\

        \bottomrule 

    \end{tabular}

    \label{tab:xar_eval}

    \vspace{0.4cm}

\end{table}

\begin{table}[t!]
    \centering
    \small 
    \caption{Evaluation of FID and CLIPScore for Harmon with continuous speculative decoding on MSCOCO and MJHQ. The results show consistent performance between the standalone model and the speculative decoding setup.}
    \setlength{\tabcolsep}{8pt} 
    \renewcommand{\arraystretch}{1.1}
    \begin{tabular}{cc|cc|c}
        \toprule
        \multirow{2}{*}{$M_p$} & \multirow{2}{*}{$M_q$} & \multicolumn{2}{c|}{FID} & CLIPScore \\
        \cmidrule(lr){3-4} \cmidrule(lr){5-5}
        & & MSCOCO & MJHQ & MSCOCO \\
        \midrule
        \multicolumn{2}{c|}{Harmon-H} & 8.39 & 5.15 & 34.8 \\
        Harmon-H & Harmon-B & 8.38 & 5.13 & 34.7 \\
        \bottomrule
    \end{tabular}
    \label{tab:harmon_combined_eval}
\end{table}

\begin{table}[t!]

    \scriptsize

    \centering

    \caption{Harmon Evaluation of Geneval. The results demonstrate the performance across various fine-grained tasks, comparing the standalone Harmon-H model with the Harmon-H and Harmon-B combination.}

    \renewcommand{\arraystretch}{1.1}

    \begin{tabular}{cc|ccccccc} 

        \toprule 

        $M_p$ & $M_q$ & Single Obj. & Two Obj. & Counting & Colors & Position & Color Attri. & Overall \\

        \midrule 

        \multicolumn{2}{c|}{Harmon-H} & 0.99 & 0.86 & 0.66 & 0.85 & 0.74 & 0.48 & 0.76 \\

        Harmon-H & Harmon-B & 0.99 & 0.83 & 0.66 & 0.86 & 0.74 & 0.44 & 0.75 \\

        \bottomrule 

    \end{tabular}

    \label{tab:geneval_eval}

\end{table}

%% file: table/acceptance_and_prefill.tex
\begin{table}[t]
    \centering
    \begin{minipage}{0.49\linewidth} 
        \caption{Ablation study on the acceptance rate $\alpha$ and the average distance of each draft and target token with and without denoising trajectory alignment under different draft length $\gamma$.}
        \resizebox{\linewidth}{!}{
        \begin{tabular}{cc|c|cc|cc}
            \toprule
            \multirow{2}*{$M_p$} & \multirow{2}*{$M_q$} & \multirow{2}*{$\gamma$} & \multicolumn{2}{c|}{$\alpha$} & \multicolumn{2}{c}{$\mathbb{E}\left[\left\|x^q - x^p\right\|^2\right]$} \\
            \cmidrule(lr){4-5} \cmidrule(lr){6-7}
                                 &                      &                         &   w/o align & w/ align & w/o align & w/ align \\
            \midrule
            MAR-\scriptsize H & MAR-\scriptsize B & 32 & 0.07 &\textbf{0.30} & 2.56 & \textbf{1.13} \\
            MAR-\scriptsize H & MAR-\scriptsize B & 16 &0.07 &\textbf{0.33} & 2.36 & \textbf{0.91} \\
            MAR-\scriptsize H & MAR-\scriptsize B & 8 & 0.13 &\textbf{0.31} & 2.22 & \textbf{0.82} \\
            MAR-\scriptsize H & MAR-\scriptsize B & 4 & 0.14 &\textbf{0.32} & 2.17 & \textbf{0.80} \\
            \bottomrule
        \end{tabular}
        }
        \label{tab:acceptance}
    \end{minipage}%
    \hfill
    \begin{minipage}{0.48\linewidth} 
        \caption{Ablation study on pre-filling ratio. \underline{Underline} indicates the highest speedup. \textbf{Bold} means the highest $\alpha$. }
        \setlength{\tabcolsep}{1.5pt} 
        \renewcommand{\arraystretch}{1.5} 
        \resizebox{\linewidth}{!}{
        \begin{tabular}{cc|c|ccc}
            \toprule 
            \multirow{2}*{$M_p$} & \multirow{2}*{$M_q$} & \multirow{2}*{$\gamma$} & \multicolumn{3}{c}{$\alpha$/Speed} \\
            \cmidrule(lr){4-6}
                                 &                      &                         & $0\%$ & $5\%$& $15\%$   \\
            \midrule 
            MAR-\scriptsize H & MAR-\scriptsize B & 32 & 0.25/\underline{1.63}$\times$ & 0.30/\underline{1.63}$\times$ & \textbf{0.33}/1.61$\times$ \\
            MAR-\scriptsize H & MAR-\scriptsize B & 16 &0.32/1.53$\times$&0.33/\underline{1.52}$\times$ & \textbf{0.34}/1.51$\times$ \\
            MAR-\scriptsize H & MAR-\scriptsize B & 8 &0.33/\underline{1.47}$\times$& 0.31/\underline{1.47}$\times$ & \textbf{0.34}/1.44$\times$ \\
            MAR-\scriptsize H & MAR-\scriptsize B & 4 & 0.31/\underline{1.21}$\times$ & 0.32/\underline{1.21}$\times$ & \textbf{0.34}/1.20$\times$ \\
            \bottomrule 
        \end{tabular}
        }
        \label{tab:prefilling}
    \end{minipage}
    \vspace{0.2cm}
\end{table}

%% file: figure_tex/draft_vis_and_curve.tex
\begin{figure}[t!] 
    \centering 
    \begin{minipage}[t]{0.54\textwidth} 
        \centering 
    \includegraphics[width=1.0\linewidth]{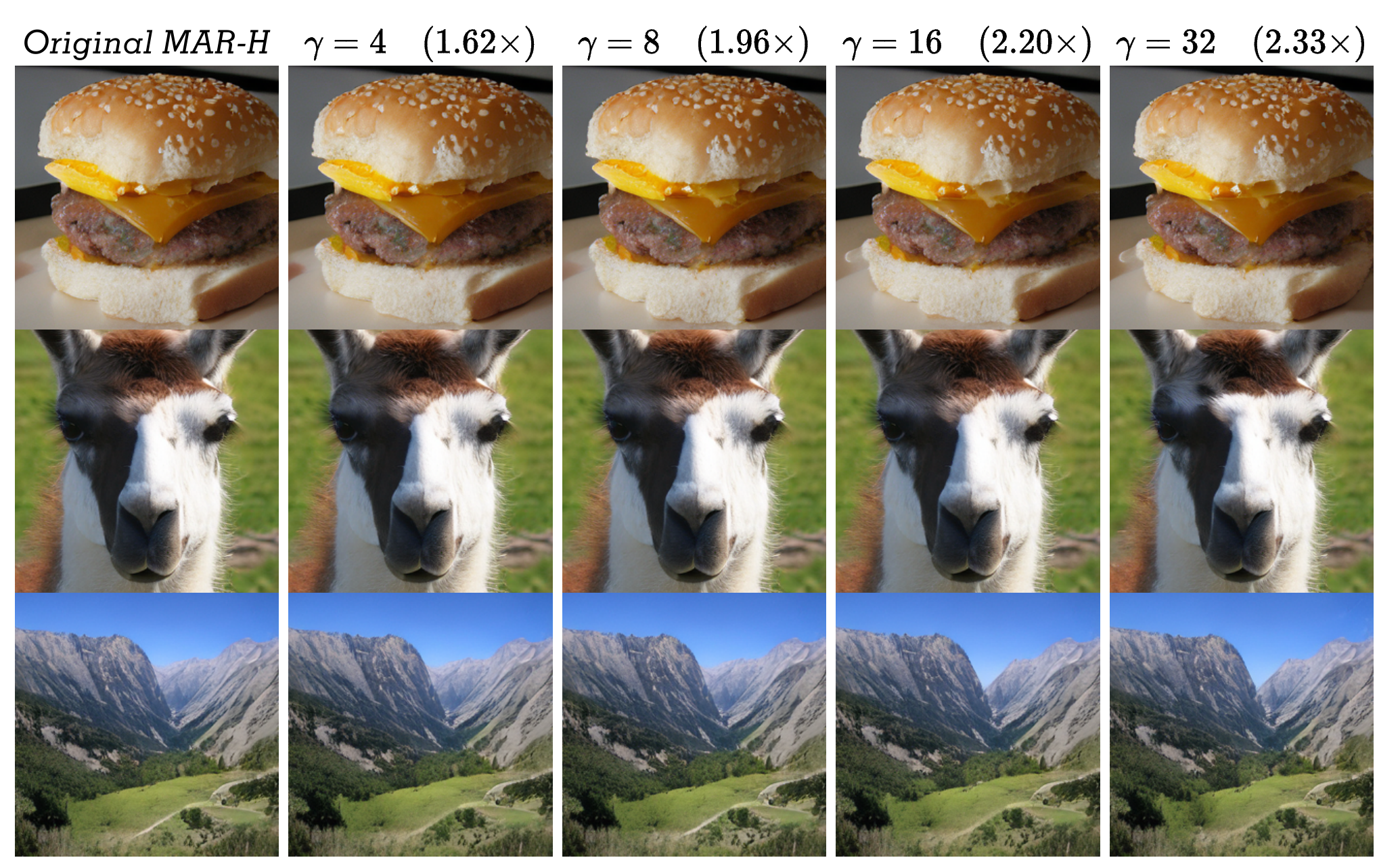}
    \caption{
    Qualitative Comparison Results. We show the images using different draft length $\gamma$.
    } 
    \label{fig:figure_comparasion}
    \end{minipage}
    \hfill
    \begin{minipage}[t]{0.44\textwidth}
        \centering
        \includegraphics[width=1.0\linewidth]{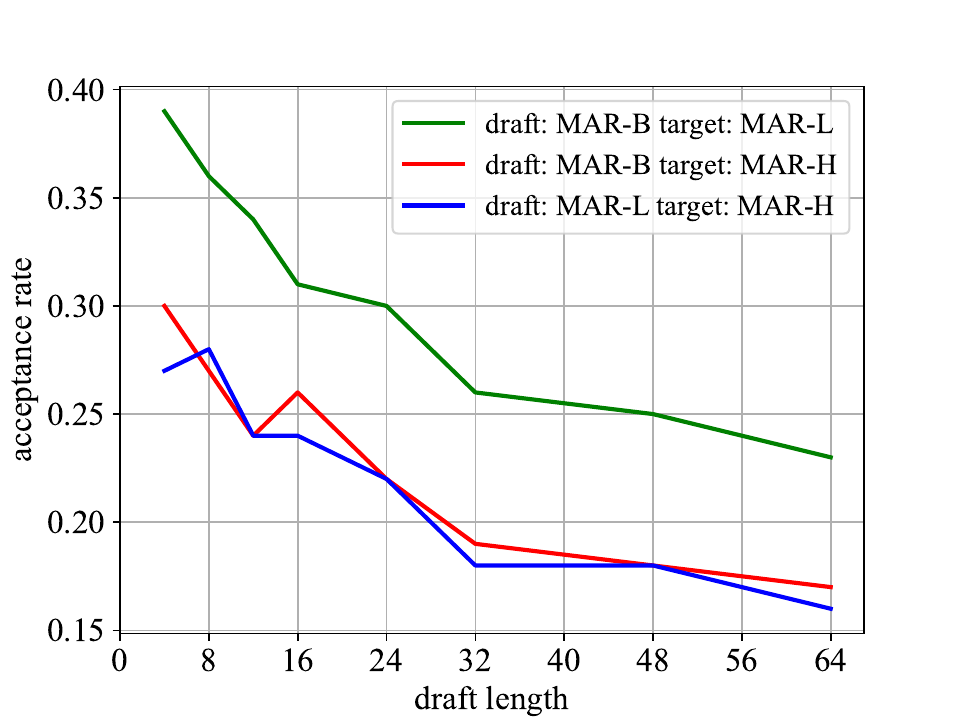}
    \caption{
    Plots of acceptance rate $\alpha$ along with draft lengths $\gamma$ on MAR.
    } 
    \label{fig:figure_acceptance_gamma}
    \end{minipage}
\end{figure}

%% file: figure_tex/figure_prefilling.tex
\begin{figure}[h]
    \centering
    \includegraphics[width=1.0\textwidth]{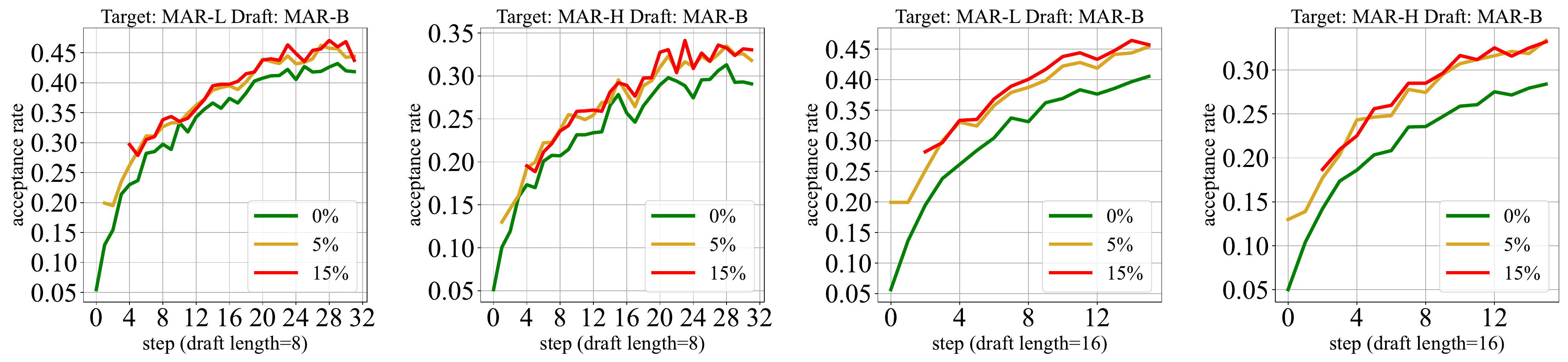}
    \caption{
    Plots of per-step acceptance rate $\alpha$ under different pre-filling ratios, along with different draft length $\gamma$, averaged on 1000 samples from MAR.}
    \label{fig:figure_prefilling}
\end{figure}

%% file: figure_tex/figure_align_vis_and_prefill_vis.tex
\begin{figure}[t!] 
    \centering 
    
    \begin{minipage}[t]{0.52\textwidth}
        \centering
        \includegraphics[width=1.0\linewidth]{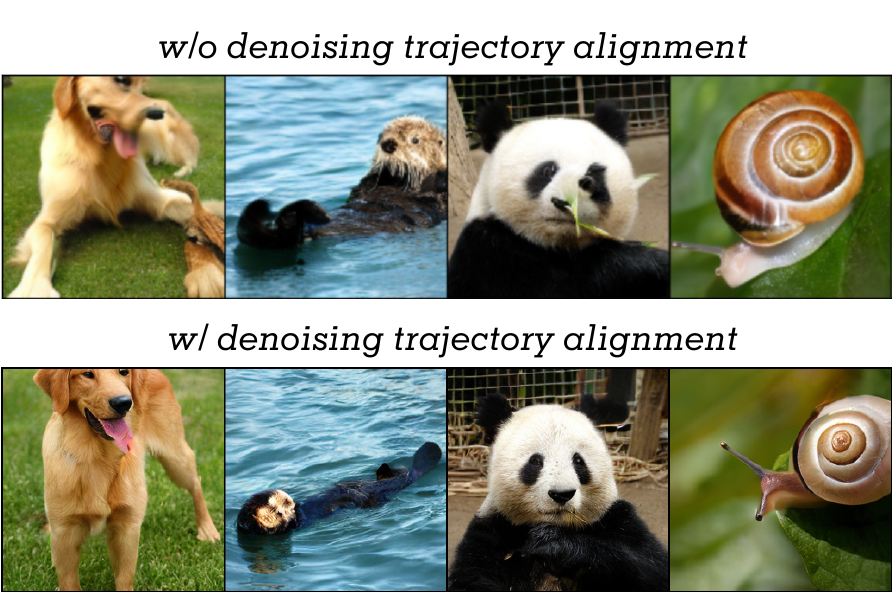}
        \caption{
        The examples without (upper) and with (lower) denoising trajectory alignment. After alignment, the generated images exhibit a reduction in deformations and artifacts, thereby achieving higher quality.
        } 
        \label{fig:figure_align_vis}
    \end{minipage}
    \hfill 
    \begin{minipage}[t]{0.46\textwidth}
        \centering
        \includegraphics[width=1.0\linewidth]{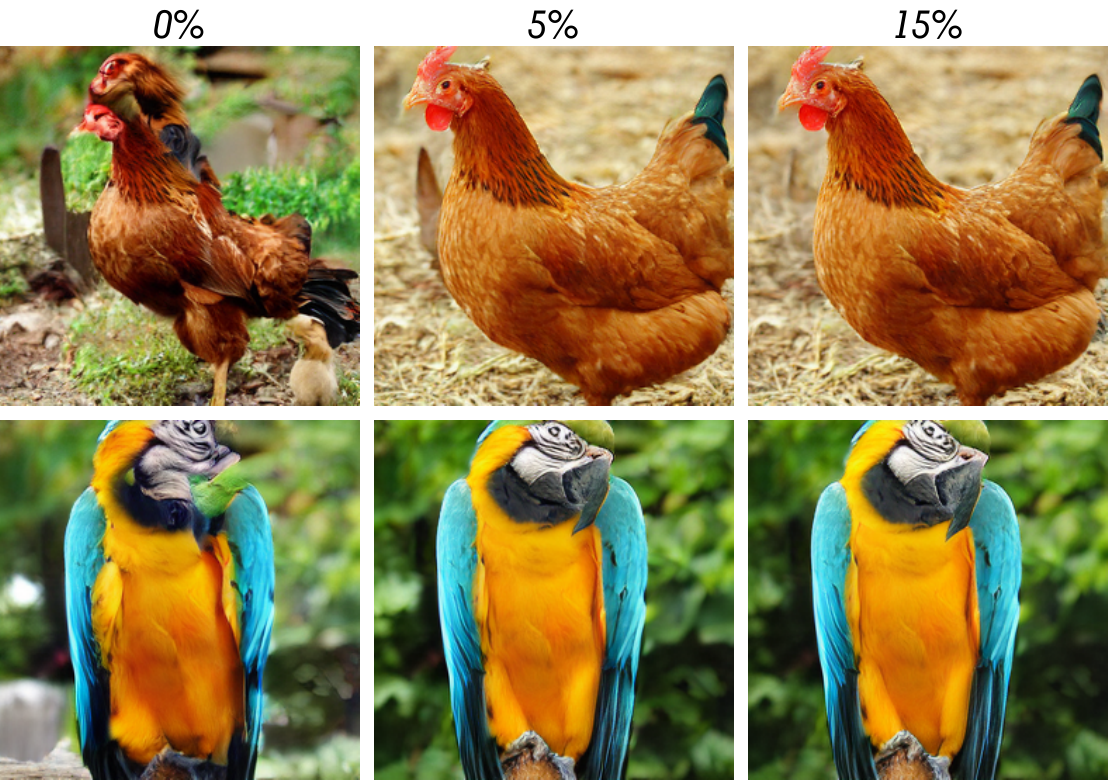}
        \caption{
        Comparing image generation quality under different token pre-filling portions. The integration of token pre-filling mitigates anomalies and synthesis errors, enhancing visual fidelity.
        } 
        \label{fig:figure_prefil_vis}
    \end{minipage}
    
\end{figure}

%% file: sec/5_conclusion.tex
\section{Conclusion}
\label{sec:conclusion}
We present continuous speculative decoding to accelerate continuous visual AR models.
We propose denoising trajectory alignment based on proximity in the reparameterization theorem and token pre-filling to enhance the acceptance rate. Acceptance-rejection sampling is introduced with a proper upper bound to sample the modified distribution without analytic expression. The repetitive diffusion model inference is tackled by reusing denoising trajectory alignment.
Extensive experiments show that our algorithm achieves over $2\times$ speedup while maintaining the output distribution.
We expect our work will provide more thoughts and insights into the inference acceleration with continuous autoregressive models in vision and other domains.

%% file: sec/X_suppl.tex
\setcounter{section}{0}
\section*{Appendix}

\section{Detailed Proofs}\label{sec:proof}
We will provide a more detailed process and proof of continuous speculative decoding.

The output of the whole model is composed of the condition from the autoregressive model and the denoising process via the diffusion model. So it can be written as:
\begin{align}
    p(x_{N,0:T}|x_1,\dots,x_{N-1})\notag
    &=p(x_{N,T})\prod_{t=1}^Tp(x_{N,t-1}|x_{N,t},x_1,\dots,x_{N-1}),
\end{align}
where $N$ denotes the number of autoregressive step, $x_1,\dots,x_{N-1}$ are tokens generated by previous steps, and $t\in[0,T]$ is diffusion timestep. For simplicity, we omit $N$ and $x_{1:N-1}$ and let $Y=x_{0:T}$ to obtain $p(Y)=p(x_T)\prod_{t=1}^Tp(x_{t-1}|x_t)$. 

\subsection{Discussion of the approximated acceptance criterion}
\subsubsection{The approximation of diffusion distribution}
\label{sec:approximation}
For a token $x$, the acceptance criterion is defined as the ratio of its probability under target distribution to the one under draft distribution, that is, $p(x)/q(x)$.
In discrete form, the probability can be directly obtained. But in continuous form, the probability is usually obtained via diffusion process~\cite{ho2020ddpm,song2020denoisingddim}. Specifically, $x=x_0$ is sampled via reverse diffusion (denoising) process $p(x_T)\prod_{t=1}^Tp(x_{t-1}|x_t)$. So the probability of the entire denoising process is:
\begin{equation}
    p(x_0)=\int_{x_{1:T}}p(x_T)\prod_{t=1}^Tp(x_{t-1}|x_T)dx_{1:T},
    \label{eqn:marginal_int}
\end{equation}

However, \Eqref{eqn:marginal_int} is analytically intractable because the product of these complex terms rarely yields a simple, closed-form analytical solution to the high-dimensional integral.

As an alternative to modeling the marginal distribution of a token $x_0$, we model the joint probability of a single fixed path $p(Y)$, where $Y = [x_0, x_1, \dots, x_T]$, to characterize the probability of each denoising trajectory. The objective of $p(x)$ is to maintain the marginal distribution, while $p(Y)$ aims to maintain the probability of the denoising trajectory itself. Both methods are effective, yet they address distribution maintenance from different perspectives.

However, using $p(Y)$ is impractical for speculative decoding because the acceptance rate is low. To validate this point, we empirically recorded the average value of different kinds of likelihood ratios over 10,000 samples with a draft length of 4, including: (i) the single-path ratio $p(Y)/q(Y)$, (ii) the two-path ratio $p(Y_p)/q(Y_q)$ without denoising trajectory alignment, and (iii) the two-path ratio $p(Y_p)/q(Y_q)$ with denoising trajectory alignment, as shown in Table~\ref{tab:marginal}.

As shown in the first row of Table~\ref{tab:likelihood_ratio}, the path-space likelihood ratio is extremely small, leading to a 0\% acceptance rate. This is because the draft model’s trajectory $Y_p$ inherently diverges from the target model’s expected trajectory. In each denoising step, samples drawn from the draft model's distribution $q$ are unlikely to fall near $\mu$ of the target distribution $p$, which results in a low single-step ratio $p/q$. As the multi-step denoising process proceeds, the overall $p(Y)/q(Y)$ becomes extremely small.

The second row compares ratios derived from independent trajectories ($Y_p$ and $Y_q$), while the final $x_0$ is generated by the draft model. This $x_0$, without alignment, is highly unlikely to fall near the target model's target distribution. However, the probability values for the other steps in these trajectories are derived from the model's own path, and thus maintain a relatively reasonable value.

The third row incorporates denoising trajectory alignment. This improvement is twofold: first, our manuscript demonstrates that the expected distance decreases; and second, our analysis shows that the correlation of $p$ and $q$ becomes 1. Consequently, the samples generated by $q$ have a high probability under $p$, resulting in an increased $p/q$.

For the above reasons, our work adopts a practical approximation for: the ratio of the joint probabilities $p(Y_p)/q(Y_q)$, where both $Y_p$ and $Y_q$ share the same $x_0$. This ratio serves as a surrogate for the intractable marginal ratio. By utilizing denoising trajectory alignment, we ensure that $Y_p$ and $Y_q$ are tightly coupled (as discussed in common issue 2), making the likelihood ratio a valid surrogate and ensuring its numerical stability in practical applications.

\subsubsection{How accurately $p(Y_p)/q(Y_q)$ approximates $p(Y)/q(Y)$}

We define the joint probability of a single denoising trajectory as $p(Y)$, where $Y$ is the sequence $[x_0,x_1,...x_T]$. The path space ratio $R=p(Y)/q(Y)$, where $Y=Y_q$, should be an unbiased estimator of $p(Y)$ to ensure that the sampling method preserves the target distribution, that is:

$$\mathbb{E}_{Y_{q}\sim q}[R]=\int q(Y_q)\frac{p(Y_q)}{q(Y_q)}dY_q=\int p(Y_q)dY_q=1.$$

However, it is very inefficient. As shown in common issue 1, $p(Y)/q(Y)$ is extremely small, because the trajectory sampled from the draft model may fall into the region where the target model assigns negligible probability mass.

To improve the efficiency, we propose a new estimator, $\tilde{R} := p(Y_p)/q(Y_q)$. $\tilde{R}$ is a biased estimator; its expectation under $q$ is not equal to 1. We analyze the difference between the expectation of $\tilde{R}$ and 1, i.e., its \textbf{bias}:

$$B = \mathbb{E}_{q} [\tilde{R}] - 1 = \mathbb{E}_{q} \left[ \frac{p(Y_p)}{q(Y_q)} \right] - \mathbb{E}_{q} \left[ \frac{p(Y_q)}{q(Y_q)} \right] = \mathbb{E}_{q} \left[ \frac{p(Y_p) - p(Y_q)}{q(Y_q)} \right].$$

We perform a first-order Taylor expansion of $p(Y_p)$ at $Y_q$. Equivalently:

$$p(Y_p) = p(Y_q + (Y_p-Y_q)) \approx p(Y_q) + \nabla p(Y_q)^T (Y_p-Y_q).$$

Substitute into $B$:

$$B \approx \mathbb{E}_{q}\left[\frac{p(Y_q) + \nabla p(Y_q)^T (Y_p-Y_q)-p(Y_q)}{q(Y_q)} \right] =\mathbb{E}_{q}\left[\frac{\nabla p(Y_q)^T (Y_p-Y_q)}{q(Y_q)} \right].$$

The magnitude of $B$ is:
\begin{equation}
    |B|\approx \left|\mathbb{E}_{q}\left[\frac{\nabla p(Y_q)^T (Y_p-Y_q)}{q(Y_q)} \right]\right| \le \mathbb{E}_{q}\left[\frac{||\nabla p(Y_q)^T||}{|q(Y_q)|} ||Y_p-Y_q|| \right].
\end{equation}

Therefore, the bound of the bias $B$ is proportional to $\mathbb{E}[||Y_p - Y_q||]$, yielding an explicit mean-square error (MSE) bound.

The reduction of expected distance is discussed in Appendix~\ref{app:dta}. The expected distance without denoising trajectory alignment is:
\begin{equation}
    \mathbb{E}\left[||x_{t-1}^q - x_{t-1}^p||^2\right]=||\mu^q_t-\mu^p_t||^2 + \text{tr}[ \Sigma^p_t + \Sigma^q_t ].
\end{equation}

The expected distance with denoising trajectory alignment is:
\begin{equation}
    \mathbb{E}_{align}\left[||x_{t-1}^q - x_{t-1}^p||^2\right]
    =||\mu^q_t-\mu^p_t||^2 + \text{tr}[ \Sigma^p_t + \Sigma^q_t -2\sqrt{\Sigma^p_t\Sigma^q_t} ].
\end{equation}
    
The two distances satisfies:
\begin{equation}
    \mathbb{E}\left[||x_{t-1}^q - x_{t-1}^p||^2\right]\ge \mathbb{E}_{align}\left[||x_{t-1}^q - x_{t-1}^p||^2\right]
\end{equation}
    
The distance over the entire trajectory satisfies:
\begin{equation}
    \mathbb{E}\left[||Y_p-Y_q||\right] \le \sqrt{\sum_{t=1}^T \mathbb{E}||x_t^q-x_t^p||^2 }.
\end{equation}

Substituting into $B$ yields the final first-order error bound:
\begin{align}
    |B|&\le \mathbb{E}_{q}\left[\frac{||\nabla p(Y_q)^T||}{q(Y_q)} ||Y_p-Y_q|| \right]\\
    &\le \mathbb{E}_{q}\left[\frac{||\nabla p(Y_q)^T||}{q(Y_q)} \sqrt{\sum_{t=1}^T ||\mu^q_t-\mu^p_t||^2 + \text{tr}[ \Sigma^p_t + \Sigma^q_t ] }\right]\\
    &|B_{align}|\le \mathbb{E}_{q}\left[\frac{||\nabla p(Y_q)^T||}{q(Y_q)} \sqrt{\sum_{t=1}^T ||\mu^q_t-\mu^p_t||^2 + \text{tr}[ \Sigma^p_t + \Sigma^q_t -2\sqrt{\Sigma^p_t\Sigma^q_t}] }\right]
\end{align}

The error bound shows that the squared drift difference $||\mu^q_t-\mu^p_t||^2$ and the covariance term $\text{tr}( \Sigma^p_t + \Sigma^q_t)$ or $\text{tr}( \Sigma^p_t + \Sigma^q_t -2\sqrt{\Sigma^p_t \Sigma^q_t})$ together determine the expected bias incurred by approximating $p(Y)/q(Y)$.

Since the use of denoising trajectory alignment produces the cross term $-2\sqrt{\Sigma^p_t \Sigma^q_t}$, the bias $|B_{align}|$ is typically smaller than $|B|$, thereby supporting a more accurate approximation of $p(Y)/q(Y)$.

\subsubsection{The extent to which $p(Y_p)/q(Y_q)$ can improve the ratio}
Let:
\begin{align}
    \log R=\log p(Y_q)-\log q(Y_q),\\
    \log \tilde{R} =\log p(Y_p)-\log q(Y_q).
\end{align}

The expected difference between the two quantities can be expressed as:
$$\Delta  l:= \mathbb{E}[\log \tilde{R}] - \mathbb{E}[\log R]= \mathbb{E}[ \log p(Y_p) ] - \mathbb{E}[ \log p(Y_q) ].$$

For each term, we have:
\begin{align}
    \mathbb{E}[ \log p(Y_p)] & = \mathbb{E}_{Y\sim p}[\log p(Y)] = -H(p),\\
    \mathbb{E}[ \log p(Y_q)] &= -H(q) - D_{KL}(q||p).
\end{align}

Substituting into $\Delta  l$ yields:
\begin{align}
    \Delta  l &= [-H(p)] - [-H(q) - D_{KL}(q||p)]\notag\\
    &=D_{KL}(q||p)+[H(q)-H(p)].
\end{align}

Empirically, the larger target model is expected to be more capable and to produce more confident (lower‑entropy) predictive distributions than the smaller draft model. Therefore, we assume the draft model’s entropy $H(q)$ is larger than the target model's entropy $H(p)$. We have:
\begin{equation}
    \Delta  l =D_{KL}(q||p)+[H(q)-H(p)]\ge 0.
\end{equation}

It indicates that in the log domain, $\tilde{R}$ exceeds $R$ by $\Delta  l$. This implies that our method increases the expected log-ratio, $\mathbb{E}[\log \tilde{R}] - \mathbb{E}[\log R]=\Delta  l$, which subsequently leads to a higher expectation of the ratio, $\mathbb{E}[ \tilde{R}]$. This explains the empirically observed higher likelihood ratios and increased acceptance rates.

\subsection{Denoising Trajectory Alignment}\label{app:dta}
We obtain $x=x_0$ through the denoising process:
\begin{equation}
    p(x_{0:T})=p(x_T)\prod_{t=1}^T p(x_{t-1}|x_t),
\end{equation}
with the conditioned probability distributions as Gaussian approximated by a neural network $\theta$:
\begin{equation}
    p_\theta(x_{t-1}|x_t)=\mathcal{N}(x_{t-1};\mu_\theta(x_t,t),\Sigma_\theta(x_t,t)).
\end{equation}
Therefore, $p_\theta(x_{t-1}|x_t)$ can be computed using the PDF of the Gaussian distribution. The computation and corresponding notation of $q_\theta(x_{t-1}|x_t)$ are the same.

Empirically, $x_{t-1}$ is obtained by sampling the Gaussian distribution on the right-hand side by \textbf{reparameterization}. That is, we first sample $\varepsilon_t \sim \mathcal{N}(0,\text{I})$, and then we obtain the result by scale and shift $x_{t-1} = \sqrt{\Sigma_\theta(x_t,t)}\cdot\varepsilon_t +\mu_\theta(x_t,t)$. To this end, we can compute $p(x)$ and $q(x)$ to obtain the ratio $p(x)/q(x)$.

However, as described in Sec.~\ref{sec:method}, directly computing the $p(x)$ and $q(x)$ is algebraically correct but may lead to a low acceptance rate due to a distinct denoising trajectory. Thus, we employ the same $\epsilon_t$ in $p(x)$ and $q(x)$ to align their trajectory as closely as possible without affecting the denoising procedure and the results.

\paragraph{Proof of Theorem~\ref{thm:proximity}}Denoising trajectory alignment enhances consistency by reducing the expected inter-sample distance throughout the denoising process. 
Suppose $x^p_{t-1}=\sqrt{\Sigma^p_t}\cdot\varepsilon_t^p+\mu^p_t$ and $x^q_{t-1}=\sqrt{\Sigma^q_t}\cdot\varepsilon_t^q+\mu^q_t$. 

\textbf{Without alignment} ($\varepsilon^p_t\neq\varepsilon^q_t$), let:
\begin{align}
    X&=x_{t-1}^q - x_{t-1}^p\notag\\
    &=(\mu^q_t - \mu^p_t)+\left(\sqrt{\Sigma^p_t}\cdot\varepsilon_t^p-\sqrt{\Sigma^q_t}\cdot\varepsilon_t^q\right)\notag\\
    &=\mu+Y,
\end{align}
where $\mu=\mu^q_t - \mu^p_t$ and $Y=\sqrt{\Sigma^p_t}\cdot\varepsilon_t^p-\sqrt{\Sigma^q_t}\cdot\varepsilon_t^q$.
The $||X||^2$ is given by:
\begin{align}
    ||X||^2=X^TX=(\mu+Y)^T(\mu+Y)=\mu^T\mu+\mu^TY+Y^T\mu+Y^TY.
\end{align}
Therefore:
\begin{align}
\mathbb{E}\left[||X||^2\right]=\mathbb{E}\left[\mu^T\mu\right]+\mathbb{E}\left[\mu^TY\right]+\mathbb{E}\left[Y^T\mu\right]+\mathbb{E}\left[Y^TY\right].\label{eqn:unalign_exp_dis}
\end{align}

First of all, $\mathbb{E}\left[\mu^T\mu\right]=\mu^T\mu$. 

Since:
\begin{align}
    \mathbb{E}\left[Y\right]=\mathbb{E}\left[\sqrt{\Sigma^p_t}\cdot\varepsilon_t^p-\sqrt{\Sigma^q_t}\cdot\varepsilon_t^q\right]=0\quad (\text{for } \mathbb{E}\left[\varepsilon_t^p\right]=\mathbb{E}\left[\varepsilon_t^q\right]=0),
\end{align}
we have $\mathbb{E}\left[\mu^TY\right]=\mathbb{E}\left[Y^T\mu\right]=0$.

For $\mathbb{E}\left[Y^TY\right]$:
\begin{align}
    Y^TY&=(\sqrt{\Sigma^p_t}\cdot\varepsilon_t^p-\sqrt{\Sigma^q_t}\cdot\varepsilon_t^q)^T (\sqrt{\Sigma^p_t}\cdot\varepsilon_t^p-\sqrt{\Sigma^q_t}\cdot\varepsilon_t^q) \notag\\
    &= (\varepsilon_t^p)^T (\sqrt{\Sigma^p_t})^T \sqrt{\Sigma^p_t} \varepsilon_t^p + (\varepsilon_t^q)^T (\sqrt{\Sigma^q_t})^T \sqrt{\Sigma^q_t} \varepsilon_t^q \notag\\
    &- (\varepsilon_t^p)^T (\sqrt{\Sigma^p_t})^T \sqrt{\Sigma^q_t} \varepsilon_t^q - (\varepsilon_t^q)^T (\sqrt{\Sigma^q_t})^T \sqrt{\Sigma^p_t} \varepsilon_t^p.
\end{align}

Note that $(\sqrt{\Sigma^p_t})^T \sqrt{\Sigma^p_t}=\Sigma^p_t$ and $(\sqrt{\Sigma^q_t})^T \sqrt{\Sigma^q_t}=\Sigma^q_t$, we have:
\begin{align}
    Y^TY=(\varepsilon_t^p)^T {\Sigma^p_t} \varepsilon_t^p+(\varepsilon_t^q)^T {\Sigma^q_t} \varepsilon_t^q-(\varepsilon_t^p)^T C \varepsilon_t^q-(\varepsilon_t^q)^T C^T \varepsilon_t^p\quad(C=\sqrt{\Sigma^p_t}\sqrt{\Sigma^q_t}).
\end{align}

Since $\mathbb{E}\left[\varepsilon^TM\varepsilon\right]=\text{tr}[M]$ where $\varepsilon\sim\mathcal{N}(0,I_n)$ and $M$ is a matrix, and $\varepsilon_t^p$ and $\varepsilon_t^q$ are independent (expectation of the cross term is $0$), we have:
\begin{align}
    \mathbb{E}\left[Y^TY\right]&=\text{tr}[\Sigma^p_t]+\text{tr}[\Sigma^q_t]+0+0=\text{tr}[\Sigma^p_t+\Sigma^q_t].
\end{align}

Put these results to \Eqref{eqn:unalign_exp_dis} makes:
\begin{align}
    \boxed{\mathbb{E}_{\varepsilon^p_t\neq\varepsilon^q_t}\left[\left\|x_{t-1}^q - x_{t-1}^p\right\|^2\right]
    =\left\|\mu^q_t - \mu^p_t\right\|^2 + \text{tr}\left[\Sigma^q_t + \Sigma^p_t\right].}
    \label{eq:no_align_d}
\end{align}

\textbf{With alignment} ($\varepsilon_t=\varepsilon^p_t=\varepsilon^q_t$), we have:
\begin{align}
    X&=(\mu^q_t - \mu^p_t)+\left(\sqrt{\Sigma^p_t}-\sqrt{\Sigma^q_t}\right)\varepsilon_t\notag\\
    &=\mu+Y\varepsilon_t,
\end{align}
where $\mu=\mu^q_t - \mu^p_t$ and $Y=\sqrt{\Sigma^p_t}-\sqrt{\Sigma^q_t}$. We have:
\begin{align}
    ||X||^2&=(\mu+Y\varepsilon_t)^T(\mu+Y\varepsilon_t)\notag\\
    &=\mu^T\mu+\mu^TY\varepsilon_t+\varepsilon_t^TY^T\mu+\varepsilon_t^TY^TY\varepsilon_t.\label{eqn:align_exp_dis}
\end{align}

Since $\mu^TY\varepsilon_t$ and $\varepsilon_t^TY^T\mu$ are scalars, we have:
\begin{align}
    \mu^TY\varepsilon_t+\varepsilon_t^TY^T\mu=2\mu^TY\varepsilon_t.
\end{align}

So:
\begin{align}
    ||X||^2&=||\mu||^2+2\mu^TY\varepsilon_t+\varepsilon_t^TY^TY\varepsilon_t.
\end{align}

Therefore:
\begin{align}
    \mathbb{E}\left[||X||^2\right]=\mathbb{E}\left[||\mu||^2\right]+2\mathbb{E}\left[\mu^TY\varepsilon_t\right]+\mathbb{E}\left[\varepsilon_t^TY^TY\varepsilon_t\right].
\end{align}

Note that $\mathbb{E}\left[||\mu||^2\right]=||\mu||^2$, and $2\mathbb{E}\left[\mu^TY\varepsilon_t\right]=2\mu^TY\mathbb{E}\left[\varepsilon_t\right]=0$. 

Also, $\mathbb{E}\left[\varepsilon_t^TY^TY\varepsilon_t\right]=\text{tr}[Y^TY]$, we have:
\begin{align}
    \text{tr}[Y^TY]&=\text{tr}[(\sqrt{\Sigma^p_t}-\sqrt{\Sigma^q_t})^T(\sqrt{\Sigma^p_t}-\sqrt{\Sigma^q_t})]\notag\\
    &=\text{tr}[{\Sigma^p_t}]+\text{tr}[{\Sigma^q_t}]-2\text{tr}[\sqrt{\Sigma^p_t\Sigma^q_t}]\notag\\
    &=\text{tr}[{\Sigma^p_t}+{\Sigma^q_t}]-2\text{tr}[\sqrt{\Sigma^p_t\Sigma^q_t}].
\end{align}

Put these results in \Eqref{eqn:align_exp_dis} makes:
\begin{align}
    \boxed{\mathbb{E}_{\varepsilon^p_t=\varepsilon^q_t}\left[\left\|x_{t-1}^q - x_{t-1}^p\right\|^2\right] = \left\|\mu^q_t - \mu^p_t\right\|^2 + \text{tr}[{\Sigma^p_t}+{\Sigma^q_t}]-2\text{tr}[\sqrt{\Sigma^p_t\Sigma^q_t}].}
    \label{eq:align_d}
\end{align}
Subtracting \Eqref{eq:align_d} from \Eqref{eq:no_align_d} yields:
\begin{align}
    \Delta\mathbb{E}&=\mathbb{E}_{\varepsilon^p_t\neq\varepsilon^q_t}\left[\left\|x_{t-1}^q - x_{t-1}^p\right\|^2\right]-\mathbb{E}_{\varepsilon^p_t=\varepsilon^q_t}\left[\left\|x_{t-1}^q - x_{t-1}^p\right\|^2\right]\notag\\
    &=\text{tr}\left[\Sigma^q_t + \Sigma^p_t\right]-\text{tr}[{\Sigma^p_t}+{\Sigma^q_t}]+2\text{tr}[\sqrt{\Sigma^p_t\Sigma^q_t}]\notag\\
    &=2\cdot\text{tr}\left[\sqrt{\Sigma^q_t\Sigma^p_t}\right]\ge 0.
\end{align}

Furthermore, alignment also simplifies the computation of $\frac{p(x)}{q(x)}$. Note that in Gaussian distribution, we have:
\begin{align}
    p(x)&=\frac{1}{(\sqrt{2\pi})^n\sqrt{|\Sigma|}}\exp{\{\frac{1}{2}(x-\mu)^T\Sigma^{-1}(x-\mu)\}}\notag\\
    &=\frac{1}{(\sqrt{2\pi})^n\sqrt{|\Sigma|}}\exp{\{\varepsilon_t^T\varepsilon_t\}}.
\end{align}
Since we have the same $\epsilon_t$ of both $p(x)$ and $q(x)$, the exponential term can be eliminated to obtain:
\begin{align}
    \frac{p(x)}{q(x)}&=\frac{\frac{1}{(\sqrt{2\pi})^n\sqrt{|\Sigma_p|}}\exp{\{\frac{1}{2}(x-\mu_p)^T\Sigma_p^{-1}(x-\mu_p)\}}}{\frac{1}{(\sqrt{2\pi})^n\sqrt{|\Sigma_q|}}\exp{\{\frac{1}{2}(x-\mu_q)^T\Sigma_q^{-1}(x-\mu_q)\}}}\notag\\
    &=\frac{\sqrt{|\Sigma_q|}}{\sqrt{|\Sigma_p|}}.
\end{align}
Therefore, for all timesteps $t$:
\begin{align}
    \frac{p(x)}{q(x)}&=\frac{p(x_T)\prod_{t=2}^T p(x_{t-1}|x_t)}{q(x_T)\prod_{t=2}^T q(x_{t-1}|x_t)}\cdot\frac{p(x_0|x_1)}{q(x_0|x_1)}\notag\\
    &=\frac{\prod_{t=2}^T \sqrt{|\Sigma_{q,t}|}}{\prod_{t=2}^T\sqrt{|\Sigma_{p,t}|}}\cdot\frac{p(x_0|x_1)}{q(x_0|x_1)}\notag\\
    &=\Sigma\cdot\frac{p(x_0|x_1)}{q(x_0|x_1)},
\end{align}
where $\Sigma$ is the cumulative product of $\frac{\sqrt{|\Sigma_{q,t}|}}{\sqrt{|\Sigma_{p,t}|}}$ along the denoising intermediate steps. Also, since $x_0\sim q(x)$ is verified by the target model, $p(x_0|x_1)$ is not obtained by denoising. It is obtained by substituting $x_0$ into $p(x_0|x_1)$. We keep the two terms since they should be computed separately.

\subsection{Acceptance-Rejection Sampling}
After rejection, we should resample a new output from:
\begin{equation}
    p^\prime (Y)=\frac{max(0,p(Y)-q(Y))}{\int_{Y^\prime}max(0,p(Y^\prime)-q(Y^\prime))dx^\prime}.
\end{equation}
But $Z$ is hard to obtain. This integral $Z=\int_{x^\prime}max(0,p(x^\prime)-q(x^\prime))dx^\prime$ is difficult to compute and may introduce computation errors if we employ an approximation. This integral also does not have an analytical form. 

On the other hand, sampling from proposal distribution $p(Y)$ requires the diffusion loss module to forward for another time, since the entire distribution is determined by all the denoising steps. But in practice, extra model inference introduces heavy overhead and extra latency, and may reduce the speed of speculative decoding, which is harmful for the whole algorithm.

\paragraph{Proof of Corollary~\ref{thm:simplified}} The introduction of acceptance-rejection sampling can eliminate $Z$ by $M=1/Z$. The denoising trajectory alignment can simplify the expression and avoid repetitive diffusion model inference. The result is given by:
\begin{align}
    \alpha_s&=\frac{max(0,p(Y)-q(Y))/Z}{p(Y)/Z}\notag\\&=\frac{max(0,p(Y)-q(Y))}{p(Y)}\notag\\
    &=\frac{max(0,p(x_T)p_\theta(x_0|x_1^p)\prod_{t=2}^T p_\theta(x_{t-1}^p|x_t^p)-q(x_T)q_\theta(x_0|x_1^q)\prod_{t=2}^T q_\theta(x_{t-1}^q|x_t^q))}{p(x_T)p_\theta(x_0|x_1^p)\prod_{t=2}^T p_\theta(x_{t-1}^p|x_t^p)}\notag\\
    &=\frac{max(0,\Sigma\cdot p_\theta(x_0|x^p_1)-q_\theta(x_0|x^q_1))}{\Sigma\cdot p_\theta(x_0|x^p_1)}
\end{align}

Afterward, we can obtain the computable results by eliminating the intermediate denoising term denoted as $\Sigma$. The final expression can be derived easily. The modified distribution can be sampled using this approach.




\section{Algorithm}
\Algref{alg:sample_with_approx} shows this procedure of continuous speculative decoding algorithm with our implementation of denoising trajectory alignment and acceptance-rejection sampling.

\newcommand{\COMMENTLLAMA}[1]{{\color{darkgreen} $\triangleright$ {#1}}}
\begin{algorithm}
  \caption{ContinuousSpeculativeDecodingStep}
  \label{alg:sample_with_approx}
\begin{algorithmic}
  \STATE {\bfseries Inputs:} $M_p, M_q, prefix$.
  \STATE \COMMENTLLAMA{Sample $\gamma$ guesses $x_{1,\ldots,\gamma}$ from $M_q$ autoregressively.}
  \FOR{$i=1$ {\bfseries to} $\gamma$}
    \STATE $q_i(Y_q) \gets M_q(prefix + [x_1, \ldots, x_{i-1}])$
    \STATE $x_i \sim q_i(Y_q)$
  \ENDFOR
  \STATE \COMMENTLLAMA{Run $M_p$ in parallel, keep the $\epsilon_t$ the same in $M_q$}
  \STATE $p_1(Y_p), \ldots, p_{\gamma + 1}(Y_p) \gets$
  \STATE \quad \quad $M_p(prefix), \ldots, M_p(prefix + [x_1, \ldots, x_{\gamma}])$
  \STATE $\Sigma\gets\frac{\prod_{t=2}^T \sqrt{|\Sigma_{q,t}|}}{\prod_{t=2}^T\sqrt{|\Sigma_{p,t}|}}$
  \STATE \COMMENTLLAMA{Determine the number of accepted guesses $n$.}
  \STATE $r_1 \sim U(0, 1), \dots, r_\gamma \sim U(0, 1)$
  \STATE $\frac{p_i(Y_p)}{q_i(Y_q)}\gets\Sigma\cdot\frac{p_i(x|x^p_1)}{q_i(x|x^q_1)}$
  \STATE $n \gets \min(\{ i - 1 \mid 1 \le i \le \gamma, r_i > \frac{p_i(x)}{q_i(x)} \} \cup \{ \gamma \})$
  \STATE \COMMENTLLAMA{Sample the modified distribution via}
  \STATE \COMMENTLLAMA{ acceptance-rejection sampling.}
  \IF{$n < \gamma$}
    \REPEAT
        \STATE $x_t\gets p_n(x|x^p_1)$
        \STATE $\alpha_s\gets\frac{max(0,\Sigma\cdot p_n(x_t|x^p_1)-q_n(x_t|x^q_1))}{\Sigma\cdot p_n(x_t|x^p_1)}$
        \STATE $\epsilon\sim U(0, 1)$
    \UNTIL{$\epsilon\le\alpha_s$}
  \ENDIF
  \STATE \COMMENTLLAMA{Return one token from $M_p$, and $n$ tokens from $M_q$.}
  \STATE {\bfseries return} $prefix + [x_1, \ldots, x_{n}, x_t]$
\end{algorithmic}
\end{algorithm}

\section{Limitations}
\subsection{Wall-Time Improvement}
As described in \cite{leviathan2023fast}, the expected walltime improvement is assumed to be:
\begin{equation}
    \frac{1-\alpha^{\gamma+1}}{(1-\alpha)(\gamma c+1)},
\end{equation}
where $\alpha$ is the acceptance rate of draft tokens, $\gamma$ is the draft length, and $c$ is the inference time ratio between the draft and target models. However, the existing draft model and the target model do not differ significantly in scale. For example, the inference time ratio $c$ of MAR-B (208M) over MAR-H (943M) is 0.38 (bs=$128$), which is \textbf{far more larger} than the number $0.05$ or close to $0$ mentioned in \cite{leviathan2023fast}. Increasing the batch size would reduce $c$, which is why our method shows better results on large batch size.

We anticipate that our algorithm will achieve more significant runtime improvements with larger target models, like 7B, 13B, as well as smaller draft models, like 97M, 125M. This direction warrants further investigation in future research.

\section{Implementation Details}
We have conducted extensive experiments with open-sourced continuous visual autoregressive model MAR~\cite{li2024mar} and xAR~\cite{ren2025xar} on ImageNet~\cite{deng2009imagenet} $256\times 256$ generation, and unified model Harmon~\cite{wu2025harmon} on text-to-image generation. The draft model is chosen from MAR-B (208M), xAR-B (172M) and Harmon-0.5B. The target model is chosen from MAR-H (943M), xAR-H (1.1B) and Harmon-1.5B, respectively. We use official pretrained checkpoints for all models. Since original xAR model is set to predict next cell of the image, the whole image would be generated in 4 steps, we let xAR to autoregressively predict next token at each position, as described as $k=1$ setting~\cite{ren2025xar}. So the autoregressive step of all the involved draft model is set to $1$. However, default MAR models have shown significant results for bidirectional attention in MAR. When target models verify the draft tokens, each output token can be regarded as the last since they can see every previous token. For MAR and xAR, their draft and target models utilize their respective class tokens \texttt{[cls]}, which are not shared during the speculative decoding process. Their diffusion loss is not shared either. The batch size ranges in $\{1$, $8$, $128$, $256\}$. The FID and IS are computed on 50k generated images, averaged on ten runs of evaluations. For Harmon models, batch size ranges in $\{1$, $8$, $16$, $32\}$, since larger batch size leads to cuda-out-of-memory. The generation resolution includes both 256 and 512. The draft and target model use their own text embedding respectively. The generation speed is measured on a single NVIDIA A100 GPU.

\section{Border Impacts}

\subsection{Border Forms of Output Distribution}
Various forms of continuous output spaces exist in the visual AR model. For instance, GIVT~\cite{tschannen2025givt} and DiCoDe~\cite{li2024dicode} employ Gaussian Mixture Models (GMMs) as output distribution. The PDF of token $x$ in GMMs is expressed as:
\begin{equation}
    p(x|\theta)=\sum_{k=1}^K\pi_k\mathcal{N}(x|\mu_k,\Sigma_k),
\end{equation}
where $\theta$ represents the model parameters, $\pi_k$ denotes the weights of each Gaussian distribution indexed by $k$, and $\mu_k$ and $\Sigma_k$ indicate the mean and covariance of distribution $k$, respectively. Our method is compatible with GMMs by computing $p(x)$, $q(x)$, and $p^\prime(x)$. The modified distribution $p^\prime(x)$ can still be computed through acceptance-rejection sampling. In practice, considering that current GMM methods cannot achieve competitive performance to MAR and the lack of open-source weights with different model sizes, we haven't conducted related experiments.

This applicability highlights a critical insight: our method relies on an explicit expression for the output distribution to compute $p(x)$, $q(x)$, and subsequently $p^\prime(x)$. As long as the specific functional form of the distribution can be obtained, our algorithm remains universally applicable, regardless of its particular form.

\subsection{More Variants of Diffusion Model}
Other variants of diffusion samplers are still applicable to our method. In this context, we utilize DDIM~\cite{song2020denoisingddim} as an example since it can be derived from DDPM without additional training.

Theoretically, DDIM~\cite{song2020denoisingddim} models a conditioned Gaussian distribution during the reverse process, expressed as $p_\theta^{(t)}(x_{t-1}|x_t)=q_\sigma(x_{t-1}|x_t,x_0)$. Its forward (diffusion) process is characterized as a non-Markov process. The term $q_\sigma$ relies on both $x_{t-1}$ and $x_0$, where $x_0$ is predicted by the model $f_\theta$. However, this dependency does not alter the form of the output PDF. The PDF of DDIM given in~\cite{song2020denoisingddim} remains consistent with that of DDPM and is computed using the expression $p_\theta(x_T)\prod_{t=1}^T p_\theta^{(t)}(x_{t-1}|x_t)$. Notably, our algorithm requires an explicit output PDF and does not rely on other properties. In general, the denoising process proceeds through sequential sampling from the previous sample. Let the PDF of initial noise be $p(x_T)$, and let the PDF for each sampling step be $p(x_{t-1}|x_t,c_{other})$, where $c_{other}$ represents any auxiliary conditions. The final output PDF is then determined as the cumulative product of all intermediate PDFs, namely: 
\begin{equation}
    p(x_{0:T}|c_{other})=p(x_T)\prod_{t=1}^Tp(x_{t-1}|x_t,c_{other}).
\end{equation}
This expression is valid across all diffusion models, regardless of specific samplers or diffusion processes like DDIM, Rectified Flow, etc. 

Practically, the original implementation of MAR is realized through DDPM. Converting DDPM to DDIM does not require additional training procedure~\cite{song2020denoisingddim}. We adopt the DDPM to the DDIM in the diffusion loss of MAR and present the relevant results in Table~\ref{tab:ddim}. We use MAR-B as the draft model and MAR-H as the target model. The batch size is 256, and the denoising step of DDIM is set to $100$. Our method and formulas remain applicable to other variants of diffusion models and continue to demonstrate good performance.
\input{table/ddim}

\subsection{Border Implementation Domains}
Continuous speculative decoding is not confined to autoregressive image generation tasks. It can also be applied to other domains, such as autoregressive audio or video generation in continuous spaces and broader tasks and scenarios within the continuous domain. Due to the limitations of currently available models, we have not been able to identify usable models beyond the domain of autoregressive image generation. However, as an algorithm without additional training or performance loss, speculative decoding remains one of the most optimal acceleration methods. We hope that continuous speculative decoding can provide more valuable insights and ideas for researchers in various fields and stimulate further research on speculative decoding across diverse domains.

\section{Information About Use of AI Assistants}
In the preparation of this work, we employ AI assistants to assist with refining academic language. The AI tools were used solely for improving clarity, grammatical correctness, and syntactic efficiency—tasks analogous to those performed by a human editor or linter. All conceptual contributions, technical claims, and critical analysis remain the authors’ own.

\section{Additional Experiments}
\label{sec:addtional_exp}

\subsection{Effectiveness of Draft \& Verification.}
\Figref{fig:figure_reject} demonstrates the comparative generation results using a pure draft model versus the draft \& verification paradigm. During the draft \& verification process, those suboptimal token regions in the draft results are systematically identified and substituted with higher-quality tokens during verification by the target model. This approach maintains the overall compositional integrity while significantly enriching the detail and quality of the generated images.
\input{figure_tex/supp_four}

\begin{table}[t]
        \centering
        \caption{Overall runtime of target model and acceptance-rejection sampling during the speculative decoding process at different draft length $\gamma$. Batch size is set to 1.}
	\renewcommand{\arraystretch}{0.5}
        \begin{tabular}{lcccc}
            \toprule
            Draft length $\gamma$ & 4 & 8 & 16 & 32 \\
            \midrule
            Target model runtime & 64s & 56s & 53s & 51s    \\
            \midrule
            Rejection-sampling rutime & 0.0378s & 0.0189s & 0.0095s & 0.0047s    \\
            \bottomrule
        \end{tabular}
	\label{tab:reject_times}
\end{table}

\subsection{Acceptance-Rejection Sampling}
Standard acceptance-rejection sampling favors large total variance $p(x) - q(x)$ for high acceptance rate, while speculative decoding prefers small differences for better draft model alignment. Therefore, we observe the practical total inference wall-time of target MAR model during the whole speculative decoding process at batch size=1, and the wall-time and sampling steps of acceptance-rejection sampling. As show in Table~\ref{tab:reject_times}, the rejection-sampling actually only accounts for a quite small fraction of the model runtime. Overall the inference speed is improved by speculative decoding. \Figref{fig:figure_rejection_times} illustrates the relationship between the rejection times and the draft length. Empirically, acceptance-rejection sampling often requires only a few sampling steps. The runtime consumed by this sampling process is negligible compared to the overall model inference time.

\subsection{Temperature}
Temperature $\tau$ is a crucial hyperparameter during the denoising process in MAR. The temperature setting affects the consistency between the outputs of the draft and target models. \Figref{fig:figure_temperature} illustrates the impact of the temperature $\tau$ on the acceptance rate during the generation process. The number of drafts length is set to $8$. The temperature influences the PDF of the final output distribution; a lower temperature may result in a sharper distribution, while a higher temperature may lead to a flatter distribution. The ratio $p(x)/q(x)$ can be influenced based on this.

\subsection{Classifier-Free Guidance}
\Figref{fig:figure_cfg} illustrates the relationship between draft length and the acceptance rate under different CFG scales. As the CFG scale increases, there is an overall trend of decreasing acceptance rates. This trend remains consistent mainly across each draft length. This phenomenon may indicate that the inconsistency between the draft and target models may increase as class guidance strengthens, reducing the acceptance rate.

\subsection{Comparison with Masked Generation}
MAR can achieve generation acceleration by generating multiple tokens per step. However, the cost of acceleration comes with a noticeable performance loss. Speculative decoding typically provides a $2\times$ speedup, consistent with the conclusion drawn in the application of LLMs~\cite{leviathan2023fast}. The performance is well maintained, as theoretically proved. In contrast, while masked generation can achieve a higher acceleration ratio (up to $10\times$ when the number of masks reaches 32), it cannot maintain performance, as shown in Table~\ref{tab:mask_gen}. We show the speedup ratio on MAR-H with 256 batch size. As the acceleration ratio increases, the model's performance suffers significant degradation.
\input{table/mask_generation}

\subsection{Visualization of Acceptance}
We visualize the acceptance and rejection region of each position through a 2D heatmap. As shown in \Figref{fig:figure_more_accept}, dark green blocks represent accepted tokens, while light green blocks represent rejected. We observe that tokens representing backgrounds or regions with simpler textures tend to be accepted. In contrast, more detailed positions are more likely to be rejected.
\input{figure_tex/figure_more_accept}

\subsection{Failure modes}
We present a visualization of the images associated with the observed failure modes and success modes. The resulting visualization is presented in \Figref{fig:quad_images}. Subfigures (a) and (b) depict the failure modes, while subfigures (c) and (d) illustrate the success modes. Our analysis reveals that the failure modes are predominantly localized in regions exhibiting high levels of detail and intricate texture representation. Subfigures (a) and (b) are characterized by rich details and fine textures, where the limited capacity of the draft model results in generated content that is below an acceptable quality threshold. In contrast, subfigures (c) and (d) possess comparatively lower complexity in terms of detail. The two subfigures include substantial background area devoid of pronounced details. The acceptance rate is substantially elevated within these less-detailed regions.
\input{figure_tex/failure_mode}

\section{More Qualitative Results}
In \Figref{fig:figure_more_vis}, ~\ref{fig:fox}, ~\ref{fig:balloon},~\ref{fig:ice_cream} and ~\ref{fig:volcano}, we provide more additional images generated under our continuous speculative decoding by MAR compared with the target model only. While the target model has achieved satisfactory quality in generating realistic and high-fidelity images, our continuous speculative decoding can show comparable performance, similar generation results, and much faster inference speed.
\input{figure_tex/figure_more_vis}
\input{figure_tex/figure_more_more_vis}

%% file: table/ddim.tex
\begin{table}[ht]
\centering
\caption{Speed up ratio and acceptance rate on MAR with DDIM as diffusion loss under different number of draft lengths.}
\begin{tabular}{c|c|c|c|c}
\hline
draft length $\gamma$ & 4 & 8 & 16 & 32 \\
\hline
speed up ratio  & 1.58 & 1.92 & 2.15 & 2.30 \\
\hline
acceptance rate $\alpha$  & 0.26 & 0.24 & 0.21 & 0.18 \\
\hline
\end{tabular}

\label{tab:ddim}
\end{table}

%% file: figure_tex/supp_four.tex
\begin{figure}[htbp]
  \centering
  \begin{subfigure}[t]{0.56\textwidth} 
    \centering
    \includegraphics[width=1.\linewidth]{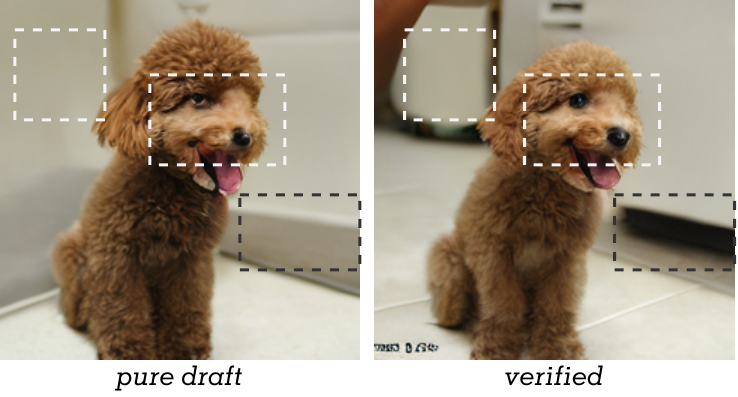}
    \caption{
    Comparison on pure draft (left) and verified (right) generation results. Regions of rejected tokens are roughly marked out.
    } 
    \label{fig:figure_reject}
  \end{subfigure}
  \hfill 
  \begin{subfigure}[t]{0.42\textwidth}
    \centering
    \includegraphics[width=1.\linewidth]{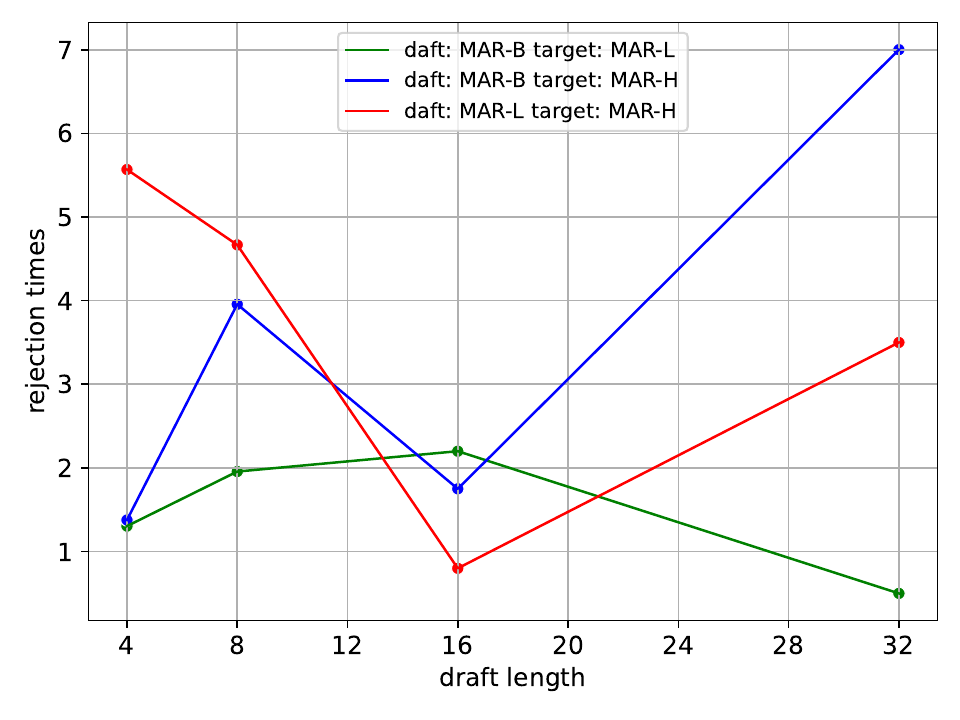}
    \caption{
    Empirical rejection times in acceptance-rejection sampling algorithm of the rejection phase.
    } 
    \label{fig:figure_rejection_times}
    
  \end{subfigure}

  \vspace{1em} 

  \begin{subfigure}[t]{0.44\textwidth}
    \centering
    \includegraphics[width=1.\linewidth]{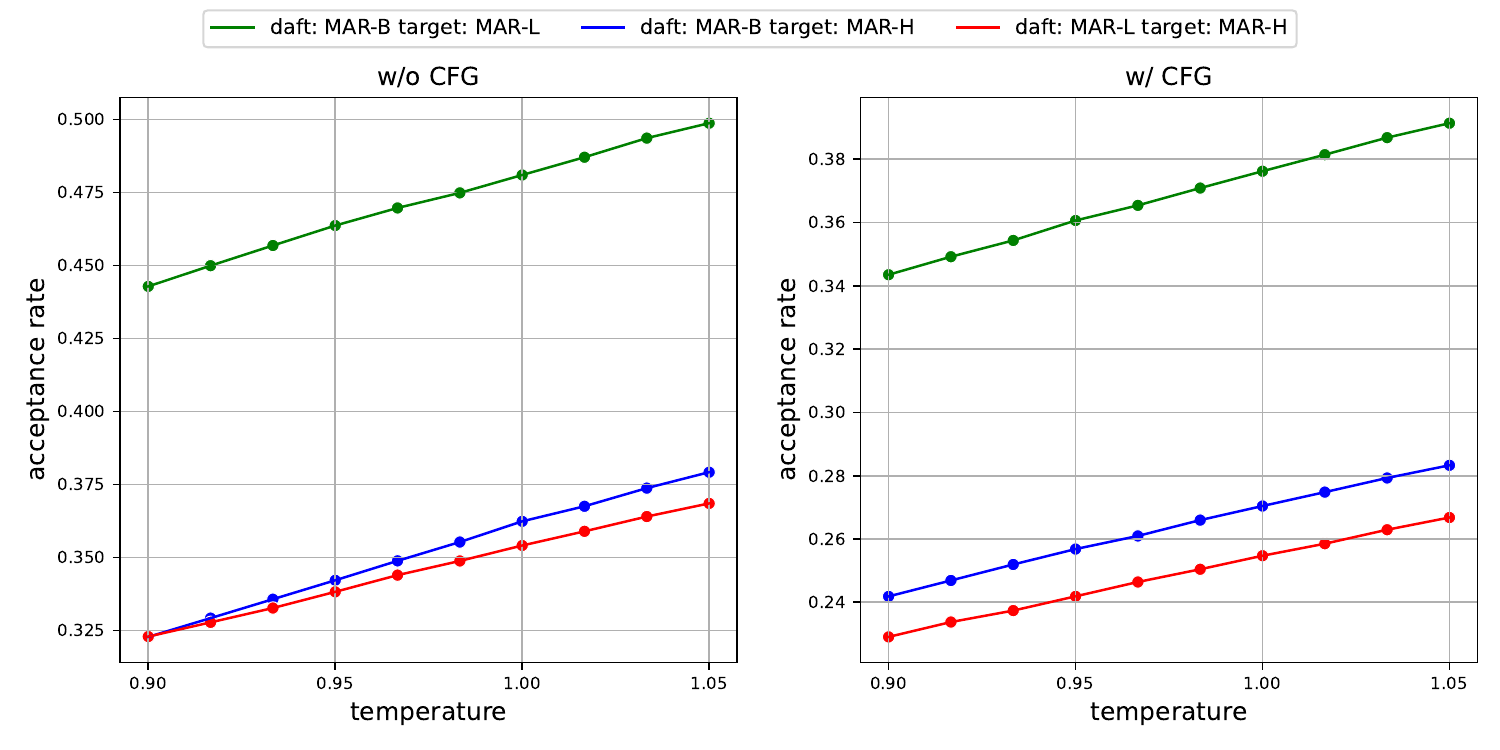}
    \caption{
    Temperature influence on acceptance rate. Left: without CFG. Right: with CFG.
    } 
    \label{fig:figure_temperature}
  \end{subfigure}
  \hfill
  \begin{subfigure}[t]{0.55\textwidth}
    \centering
    \includegraphics[width=1.\linewidth]{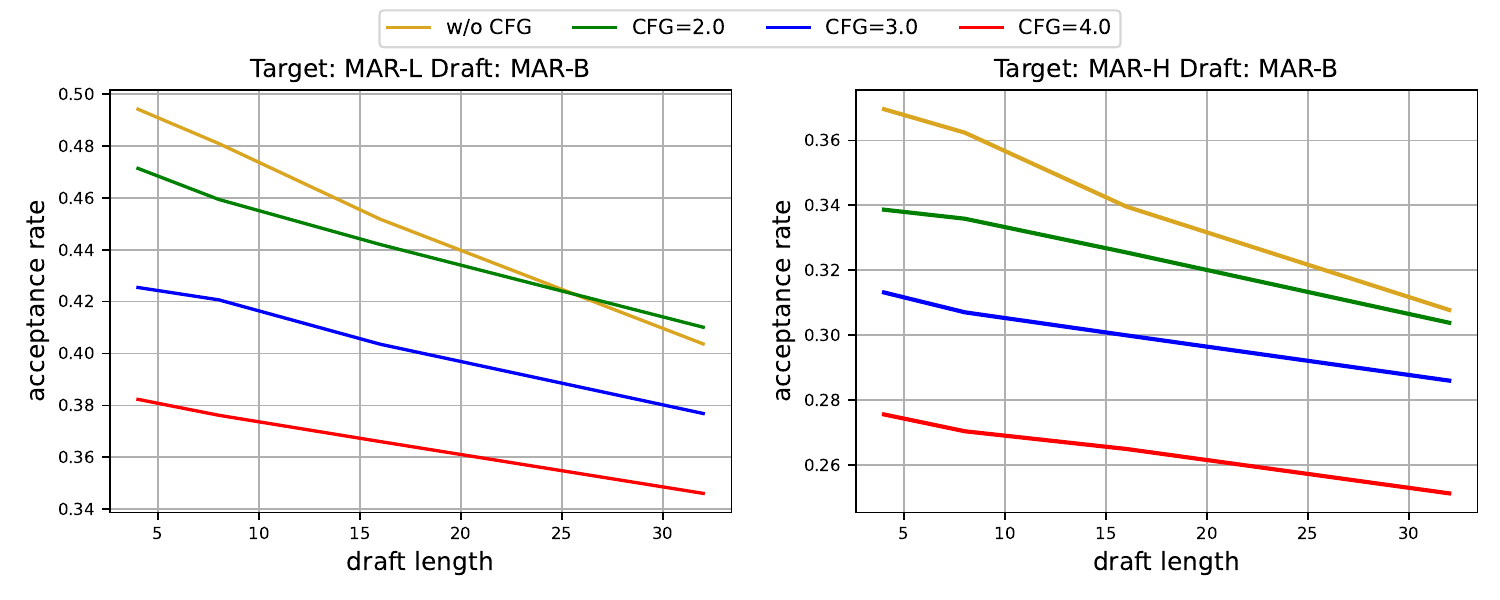}
    \caption{
    CFG scale has has a significant impact on the acceptance rate under different number of draft lengths.
    } 
    \label{fig:figure_cfg}
  \end{subfigure}

  \caption{Ablation studies on various experiment factors conducted on MAR models.} 
\end{figure}

%% file: table/mask_generation.tex
\begin{table}[h!]
\centering
\caption{Speed-up ratio and FID under different mask generation steps. Larger mask generation step can bring the model a better speedup, but it also leads to significant performance degradation.}
\setlength{\tabcolsep}{8pt} 
    \renewcommand{\arraystretch}{1.0} 
\begin{tabular}{c|cccccc}
\hline
\# mask & 2 & 4 & 8 & 16 & 32 & 64 \\
\hline
speed up & 1.99 & 3.49 & 5.54 & 7.85 & 10.02 & 11.49 \\
\hline
FID & 1.56 & 2.37 & 3.66 & 4.99 & 17.43 & 59.07 \\
\hline
\end{tabular}
\label{tab:mask_gen}
\end{table}

%% file: figure_tex/figure_more_accept.tex
\begin{figure}[h!]
    \centering
    \includegraphics[width=0.95\linewidth]{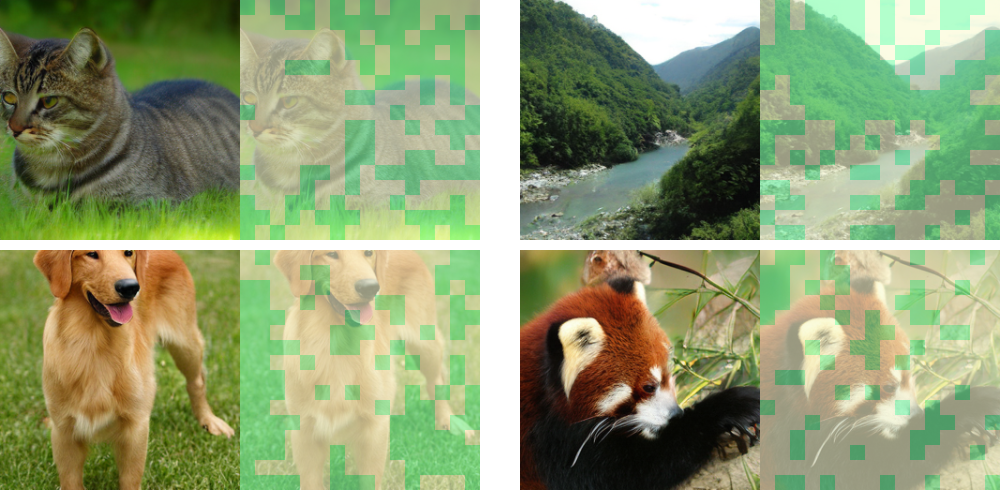}
    \caption{
    Visualizations of accepted token heatmap. Dark green: accepted. Light green: rejected.
    } 
    \label{fig:figure_more_accept}
\end{figure}

%% file: figure_tex/failure_mode.tex
\begin{figure}[h!] 
    \centering
    
    \begin{subfigure}[b]{0.24\textwidth} 
        \includegraphics[width=\linewidth, height=3cm]{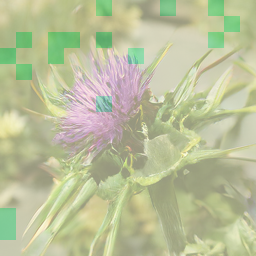} 
        \caption{$\alpha=5.88\%$} 
        \label{fig:1}
    \end{subfigure}
    \hfill 
    \begin{subfigure}[b]{0.24\textwidth}
        \includegraphics[width=\linewidth, height=3cm]{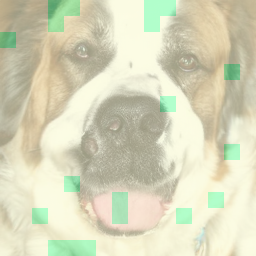} 
        \caption{$\alpha=7.06\%$}
        \label{fig:2}
    \end{subfigure}
    \hfill
    \begin{subfigure}[b]{0.24\textwidth}
        \includegraphics[width=\linewidth, height=3cm]{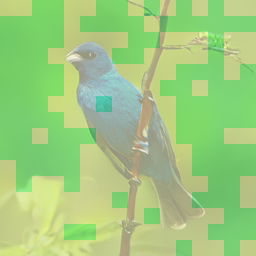} 
        \caption{$\alpha=43.53\%$}
        \label{fig:3}
    \end{subfigure}
    \hfill
    \begin{subfigure}[b]{0.24\textwidth}
        \includegraphics[width=\linewidth, height=3cm]{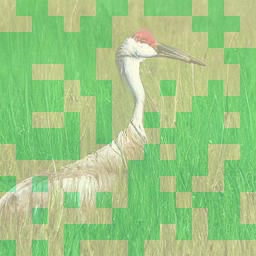} 
        \caption{$\alpha=49.41\%$}
        \label{fig:4}
    \end{subfigure}
    
    \caption{Subfigure (a) and (b): failure modes. Subfigure (c) and (d): success modes. The visualizations reveal that the failure modes are predominantly localized in regions exhibiting high levels of detail and intricate texture representation. } 
    \label{fig:quad_images}
\end{figure}

%% file: figure_tex/figure_more_vis.tex
\begin{figure*}[t!]
    \centering
    \includegraphics[width=1.0\columnwidth]{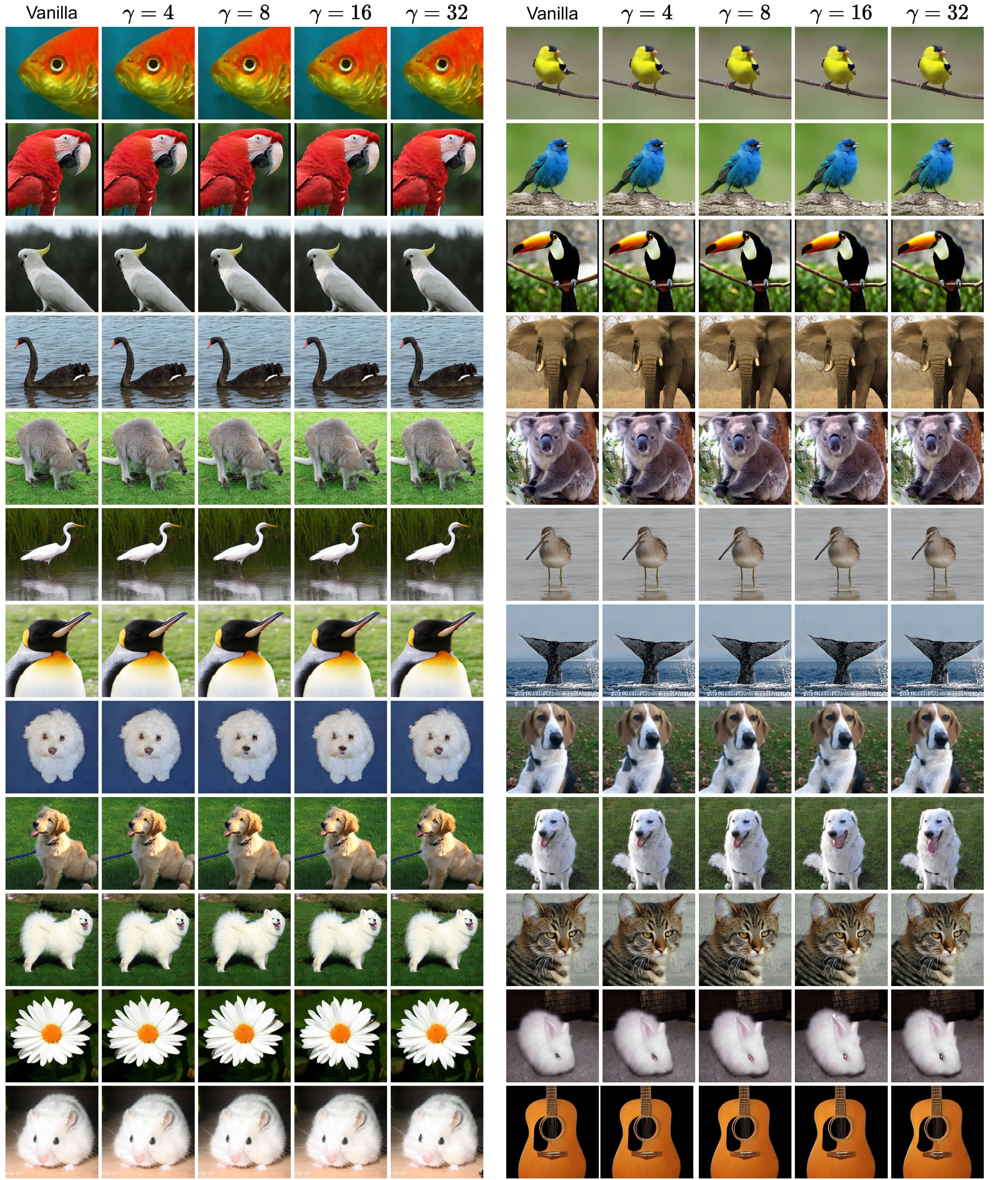}
    \caption{
    Visual quality with increasing draft length $\gamma$ compared with vanilla target model only generation. \textit{Best viewed zoom-in.}
    } 
    \label{fig:figure_more_vis}
\end{figure*}

%% file: figure_tex/figure_more_more_vis.tex
\begin{figure*}[t!]
    \centering
    \includegraphics[width=0.85\linewidth]{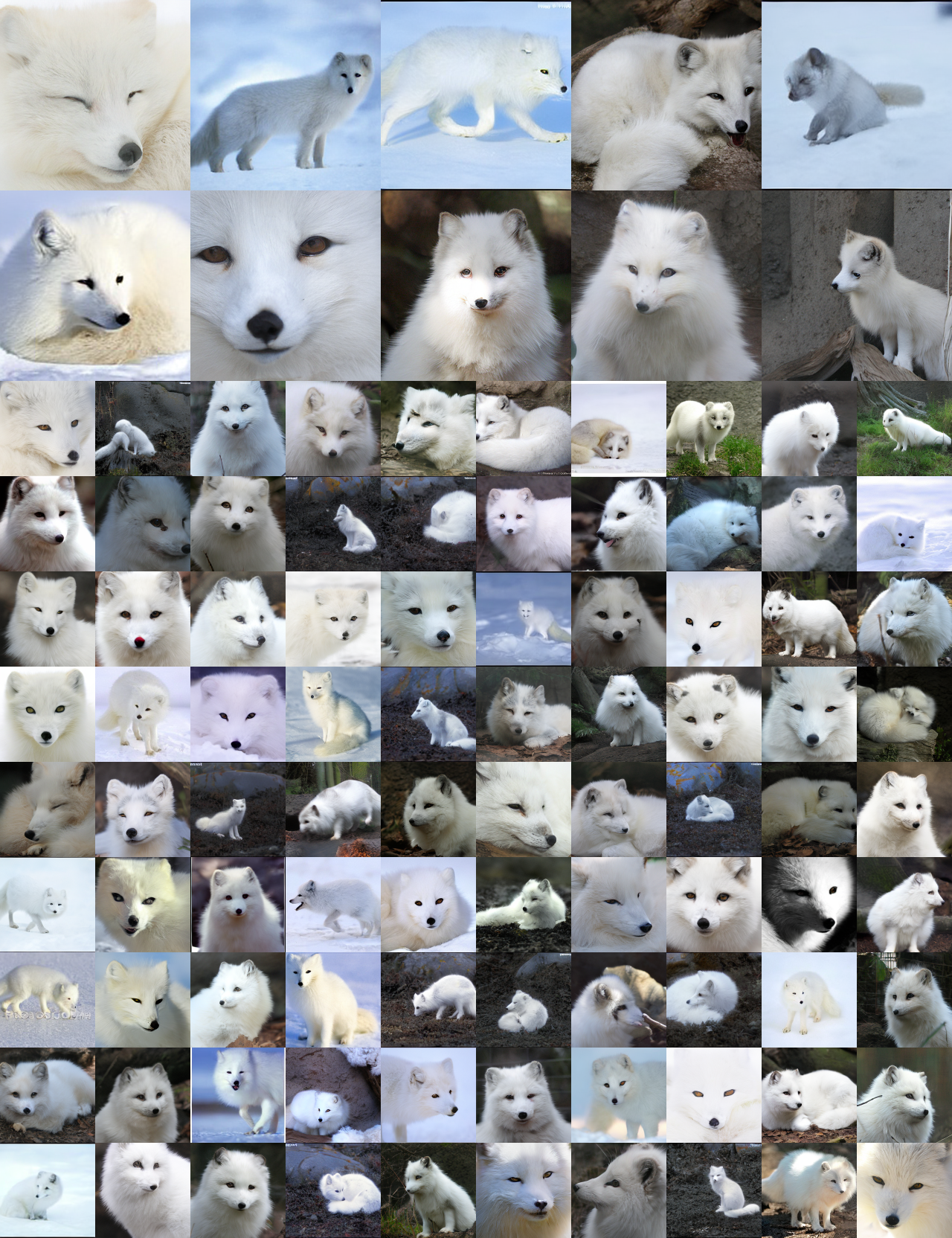}
    \caption{
    Visualization examples under $\gamma=4$. Class label: arctic fox (297).
    } 
    \label{fig:fox}
\end{figure*}

\begin{figure*}[t!]
    \centering
    \includegraphics[width=0.85\linewidth]{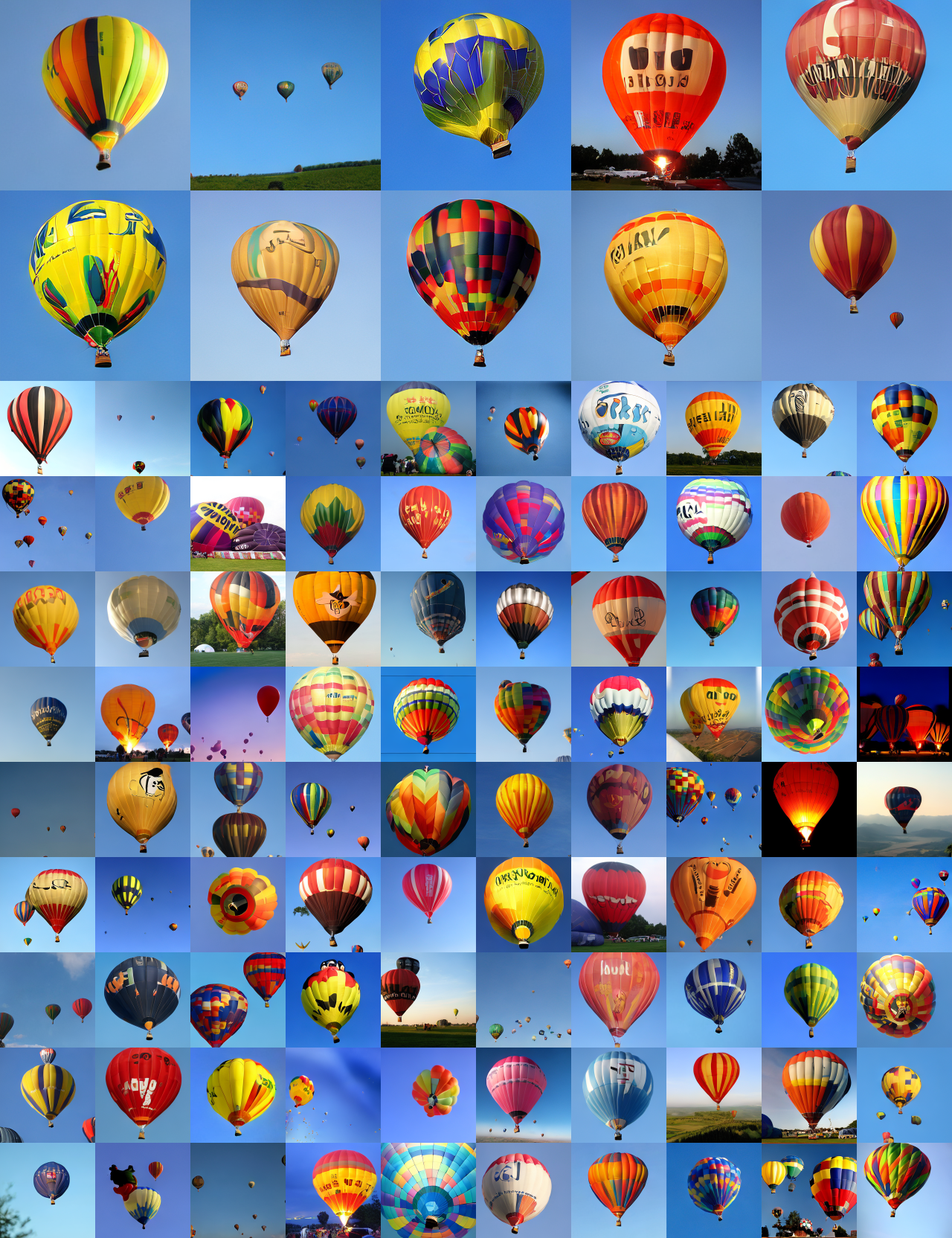}
    \caption{
    Visualization examples under $\gamma=8$. Class label: balloon (417).
    } 
    \label{fig:balloon}
\end{figure*}

\begin{figure*}[t!]
    \centering
    \includegraphics[width=0.85\linewidth]{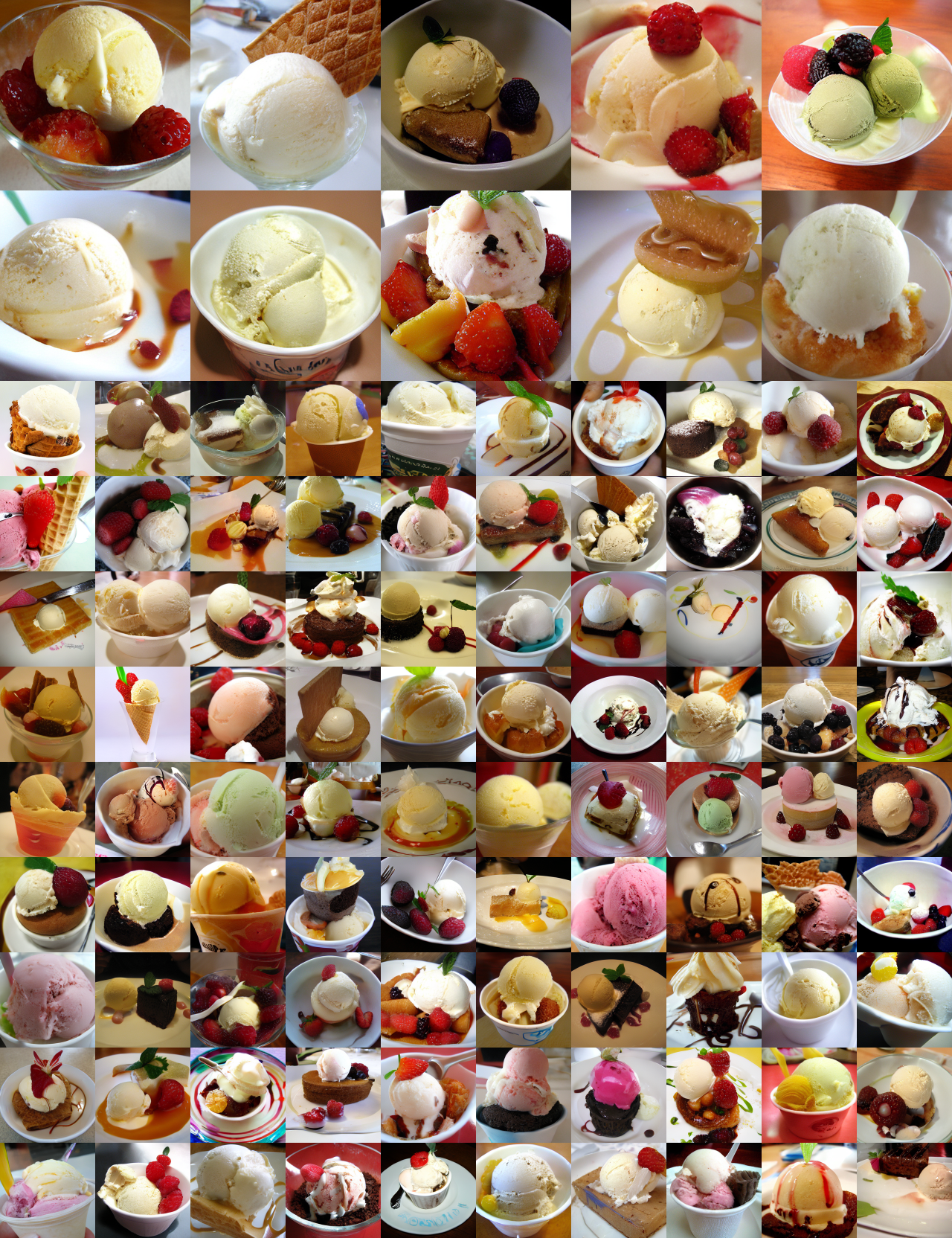}
    \caption{
    Visualization examples under $\gamma=16$. Class label: ice cream (928).
    } 
    \label{fig:ice_cream}
\end{figure*}

\begin{figure*}[t!]
    \centering
    \includegraphics[width=0.85\linewidth]{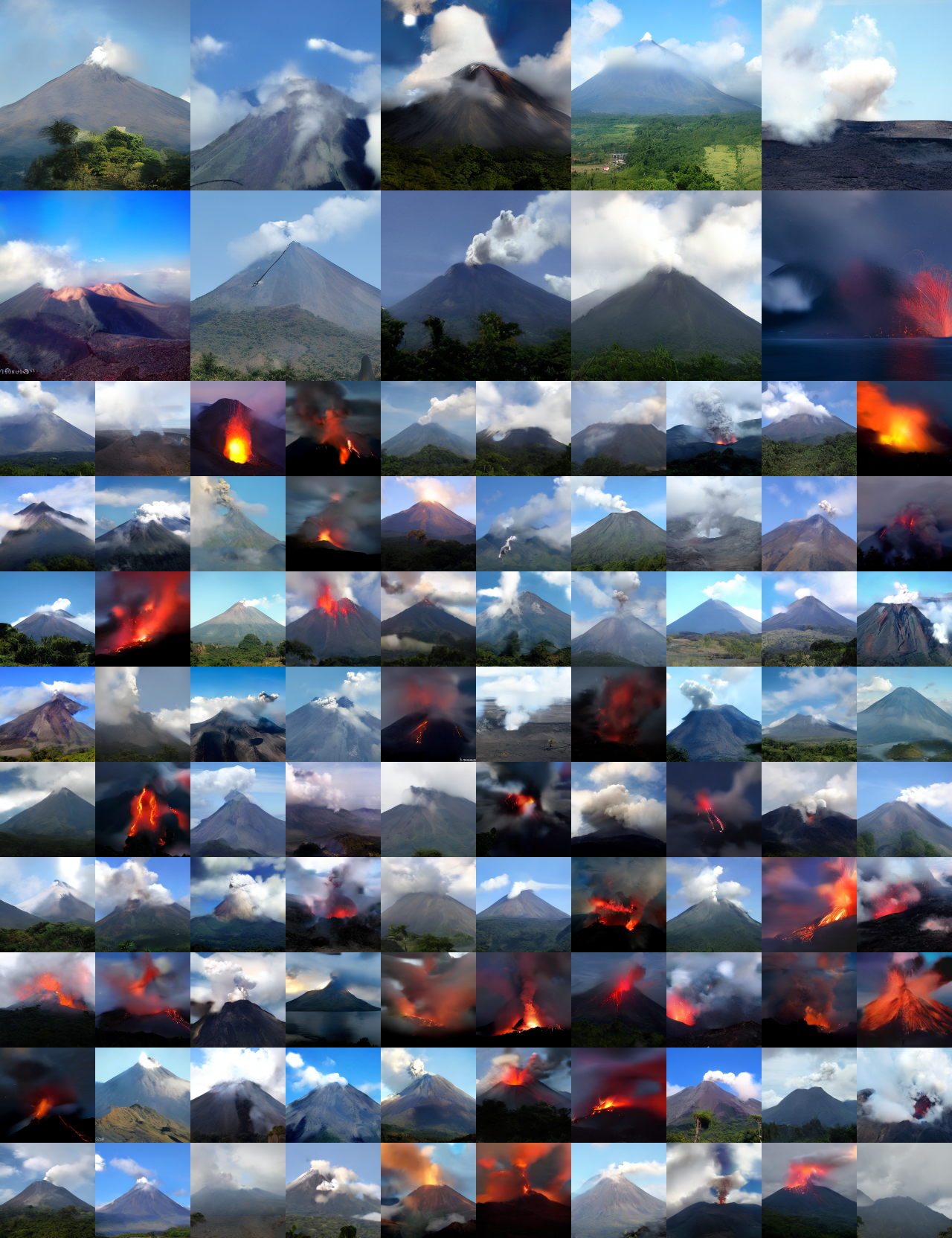}
    \caption{
    Visualization examples under $\gamma=32$. Class label: volcano (980).
    } 
    \label{fig:volcano}
\end{figure*}